%% file: main.tex
\documentclass{article} 
\usepackage{main,times}

\input{math_commands.tex}

\definecolor{citecolor}{HTML}{103C5B}
\definecolor{linkcolor}{rgb}{0.956,0.298,0.235} 
\usepackage[hidelinks,colorlinks,linkcolor=blue,citecolor=blue]{hyperref}
\usepackage{url}
\usepackage{booktabs}       
\usepackage{amsfonts}       
\usepackage{wrapfig}
\usepackage{graphicx}      
\usepackage{booktabs}
\usepackage{multirow}
\usepackage{amsmath}        
\usepackage{nicefrac}       
\usepackage{microtype}      
\usepackage{xcolor}         
\usepackage{colortbl}

\title{HieraTok: Multi-Scale Visual Tokenizer Improves Image Reconstruction and Generation}

\author{Cong Chen$^{1,2}$\thanks{ CC and ZYH contributed equally. CS is the corresponding author.}
~~
\textbf{Ziyuan Huang}$^{2*}$, 
~~
\textbf{Cheng Zou}$^{2}$,
~~
\textbf{Muzhi Zhu}$^{1,2}$, 
~~
Kaixiang Ji$^{2}$ \\
\textbf{Jiajia Liu}$^{2}$,   
~~~
\textbf{Jingdong Chen}$^{2}$,  
~~~
\textbf{Hao Chen$^{1}$, 
~~~
Chunhua Shen$^{1,2,3}$}\vspace{0.3cm} \\
$^1$ Zhejiang University \quad
$^2$ Ant Group \quad
$^3$ Zhejiang University of Technology 
}

\iclrfinalcopy 
\begin{document}

\maketitle

\begin{abstract}
In this work, we present HieraTok, a novel multi-scale Vision Transformer (ViT)-based tokenizer that overcomes the inherent limitation of modeling single-scale representations. This is realized through two key designs: (1) multi-scale downsampling applied to the token map generated by the tokenizer encoder, producing a sequence of multi-scale tokens, and (2) a scale-causal attention mechanism that enables the progressive flow of information from low-resolution global semantic features to high-resolution structural details. Coupling these designs, HieraTok achieves significant improvements in both image reconstruction and generation tasks. Under identical settings, the multi-scale visual tokenizer outperforms its single-scale counterpart by a 27.2\% improvement in rFID ($1.47 \rightarrow 1.07$). When integrated into downstream generation frameworks, it achieves a $1.38\times$ faster convergence rate and an 18.9\% boost in gFID ($16.4 \rightarrow 13.3$), which may be attributed to the smoother and more uniformly distributed latent space. Furthermore, by scaling up the tokenizer's training, we demonstrate its potential by a sota rFID of 0.45 and a gFID of 1.82 among ViT tokenizers. To the best of our knowledge, we are the first to introduce multi-scale ViT-based tokenizer in image reconstruction and image generation. We hope our findings and designs advance the ViT-based tokenizers in visual generation tasks.
\end{abstract}

\section{Introduction}
The rapid advancement of latent generative models ~\citep{ldm,Imagen,fluid,zero_shot,text_to_image} has revolutionized visual content creation, achieving remarkable success in high-quality image synthesis through architectures like diffusion models~\citep{dit,ldm,ddpm,flowmodel} and autoregressive transformers~\citep{vqgan,attention}. These tasks typically follow a two-stage training paradigm: first, a visual tokenizer~\citep{ae,vae,mae} is trained to compress images into latent representations; second, a generative model learns to map conditions to the latent distribution. While current research predominantly focuses on developing novel generative paradigms~\citep{mar,var} and scaling up generative models~\citep{gpt-4o, gemini,simplear}, we emphasize that the design of the visual tokenizer is equally crucial. The reconstruction quality of the visual tokenizer fundamentally determines the upper bound of image generation quality, while the regularity of the latent space~\citep{diffusability, vavae} governs the convergence speed of downstream generative models.

A key challenge in designing an efficient visual tokenizer is its ability to capture the hierarchical nature of visual information. Natural images inherently exhibit scale consistency, spanning from global semantics to local details. As a result, visual tasks have consistently benifited from multi-scale designs ~\citep{unet,swintransformer,chain-of-sight,mvit} , as evidenced by the success of FPN~\citep{FPN} in object detection and segmentation, as well as multi-scale autoregressive models like VAR~\citep{var} and FlowAR~\citep{flowar} in generation tasks. 


Convolutional architectures naturally model multi-scale representations by progressively reducing spatial dimensions and increasing channel depth. However, for Vision Transformer-based tokenizers~\citep{scaling_vit,gigatok,vit}, they are inherently limited to modeling single-scale representations, and no multi-scale ViT Tokenizer have been proposed so far. We attribute this limitation to the following challenges:

\begin{figure}[t]
  \centering
  \includegraphics[width=1.0\linewidth]{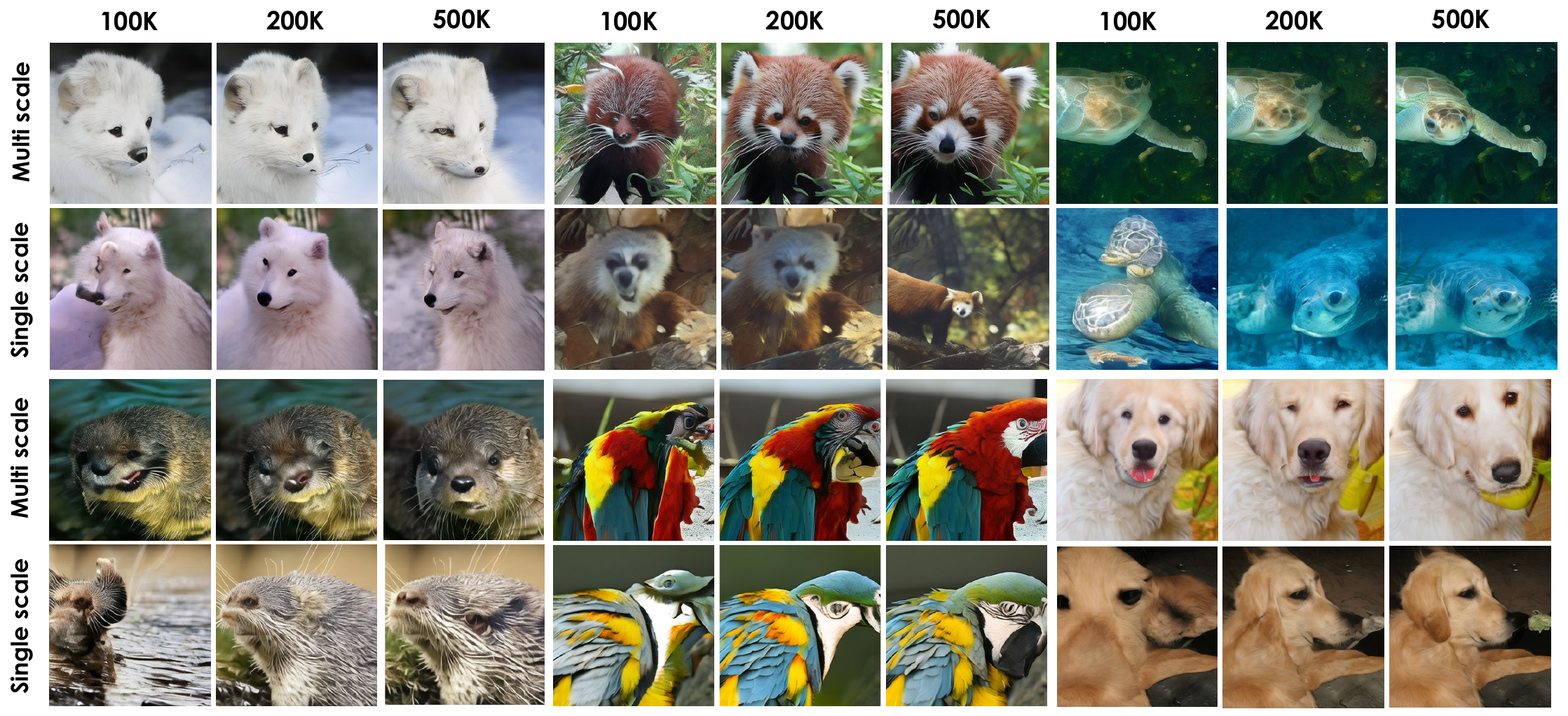}
  \caption{Comparative visualization of generation quality: multi-scale vs. single-scale tokenizers across increasing generative model training steps. (a) Top row: Samples generated by multi-scale tokenizer. (b) Bottom row: Corresponding outputs from the single-scale baseline. We train these models using the DiT-XL framework and conduct inference under identical conditions. }
\end{figure}

(1) \textbf{Fixed-length token sequences}: Convolutional networks adjust feature map sizes through convolution operations. In contrast, for ViT-based models, after patchfying the input image to a token map of a specific resolution, the number of tokens remains fixed throughout the network. This inherently restricts the model to representing fixed-length, single-scale features.

(2) \textbf{Global attention mechanism in Transformers}: While the global attention~\citep{attention} mechanism in transformers enables comprehensive token interactions, its uniform treatment of all tokens (despite positional encoding) presents difficulties in establishing hierarchical relationships. The self-attention mechanism inherently lacks inductive biases for modeling resolution-dependent feature hierarchies, making it challenging to naturally maintain the progression from high-level semantic representations to low-level structural details that is characteristic of convolutional architectures. 

Building on these insights, we introduce HieraTok, a novel and efficient multi-scale tokenizer. To overcome the fixed token sequence length in ViT, we generate a sequence of multi-scale token maps by downsampling the single-scale token map produced by the encoder. During the transformer's processing of these multi-scale token maps, we implement a scale-causal attention mechanism, which restricts high-resolution tokens to interact exclusively with preceding low-resolution tokens. This design ensures a progressive flow of information from coarse-grained semantic tokens to fine-grained structural tokens, effectively establishing a hierarchical transition from global to fine-grained local features. As seen in the Figure \ref{fig:attention}, this approach mirrors the principles of the FPN~\citep{FPN} in convolutional architectures. Furthermore, we apply corresponding interpolation to the ground truth RGB images, enabling each scale's token map to reconstruct the RGB image at its respective scale. Such supervision also embeds scale consistency into the representation space.

Our HieraTok achieves significant improvements in both reconstruction fidelity and generation efficiency. Experimental results validate that under identical training setting,the multi-scale ViT tokenizer significantly outperforms its single-scale counterpart in reconstruction tasks by a 27.2\% improvement in rFID (1.47 → 1.07). In generation tasks, our tokenizer achieves faster convergence rates in diffusion generation frameworks. Additionally, through analyses such as t-SNE~\citep{tsne} in Figure\ref{figure:tsne}, we observe that the latent space of the multi-scale ViT tokenizer is smoother and more uniformly distributed. This characteristic likely contributes to the accelerated training of generative models. These experiments confirm the effectiveness of our multi-scale tokenizer design. Furthermore, under a scaled-up experimental setting, HieraTok achieves a sota rFID of 0.45 and a highly competitive gFID of 1.82 among ViT-based tokenizers. We hope this multi-scale approach will drive further breakthroughs in image generation tasks for existing ViT-based tokenizers.

\section{Related Work}

\subsection{Tokenizers for Image Reconstruction}
Modern tokenizers compress images into continuous or discrete latent representations. Continuous tokenizers, like AE~\citep{ae, dcae, ae_intro} and VAE~\citep{vae, vae_intro, flux}, encode images into continuous feature spaces or latent distributions. In contrast, discrete tokenizers such as VQ-VAE~\citep{vqvae,vqvae2,movqgan} employ quantization to obtain discrete codes. During the first stage of generative modeling, tokenizers are optimized for high-fidelity reconstruction, where continuous representations often outperform their discrete counterparts. However, empirical observations~\citep{mgvit2, scaling_vit, vit} reveal an intriguing trade-off: improved reconstruction quality frequently correlates with degraded generation performance. Recent advances~\citep{gigatok, scaling_vit} have sought to address this limitation through better tokenizer architectures design. In this work, our multiscale tokenizer preserves structural consistency across scales, jointly improving reconstruction fidelity and generation quality.

\subsection{Image Generation Paradigm}
 Generation models learn a mapping from conditions to latent space distributions. Diffusion-based approaches (e.g., LDM~\citep{ldm}, DiT~\citep{dit}) operate in continuous spaces~\citep{flowmodel}, while autoregressive transformers (e.g., VQGAN~\citep{vqgan}) generate discrete tokens. Recently, methods like MAR~\citep{mar,fluid} have bridged these paradigms by introducing diffusion losses into autoregressive frameworks, enabling continuous autoregressive generation. Beyond architectural choices, recent work also focus on improving convergence speed. For instance, REPA~\citep{repa} accelerates generation by aligning intermediate representations with pretrained visual foundation models~\citep{clip,dino,dinov2}. Meanwhile, studies like VAVAE~\citep{vavae} and EQ-VAE~\citep{vavae} highlight the critical role of latent space structure in governing training difficulty. Our work builds on these insights by proposing a multi-scale latent space design, where explicit scale consistency yields a more generation-friendly representation~\citep{diffusability}.

\subsection{Multi-Scale visual design}
Scale invariance is a fundamental property of images. Hierarchical multi-scale features—from coarse global structures to fine local details—provide significant benefits for vision tasks. In traditional detection and segmentation, architectures like FPN~\citep{FPN}, U-Net~\citep{unet}, and later Swin-Transformer~\citep{swintransformer}, ViTDet~\citep{vitdet}, and Mask2Former~\citep{mask2former} have demonstrated the critical role of multi-scale design for fine-grained visual understanding. This principle extends to image generation: VAR~\citep{var} achieved substantial quality improvements in discrete autoregressive generation through multi-scale quantization and autoregressive strategies, while Ming-Lite~\citep{ming-lite} enhanced fine-grained details via learnable multi-scale queries. Our work identifies unique challenges in adapting multi-scale designs to ViT-based tokenizers. We address this with a simple yet effective architecture that significantly improves both reconstruction and generation quality.

\section{Method}

\begin{figure}[t]
  \centering
  \includegraphics[width=1.0\linewidth]{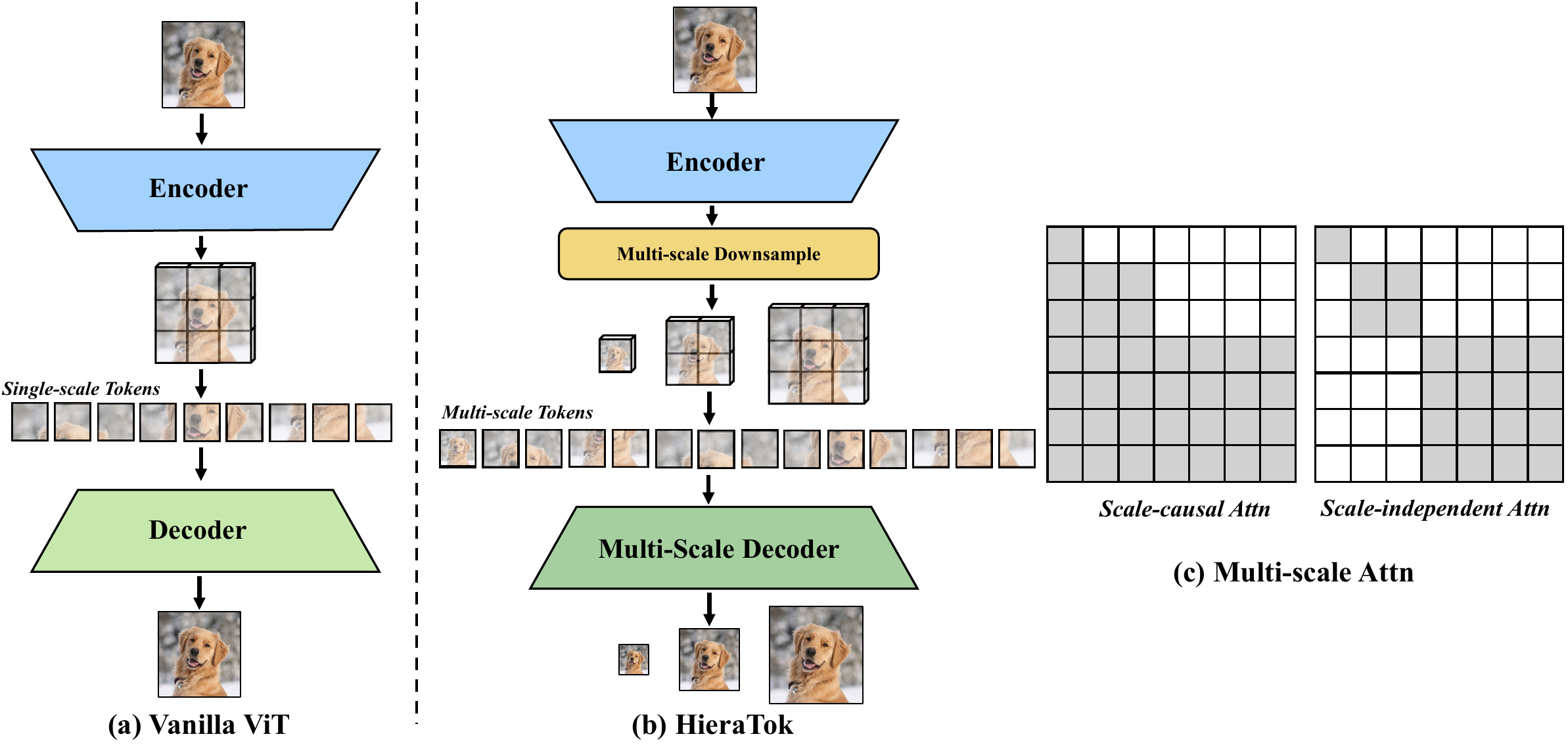}
  \caption{Different tokenizer designs: (a) Single-scale tokenizer ;
(b) Our multi-scale tokenizer, featuring two key designs of multi-scale downsampling modules and 
multi-scale attention mechanisms in the decoder;
(c) Multi-scale attention variants: Scale-causal and Scale-independent.}
\label{figure:architecture}
\end{figure}

\subsection{Preliminary: Vanilla ViT Tokenizer}
In the vanilla Vision Transformer (ViT)~\citep{vit} framework, the tokenizer encodes an input image \(\mathbf{X}\) by first applying patch embedding through convolutional or linear layers, yielding a grid of \( \frac{h}{p} \times \frac{w}{p} \times d \) tokens, where \( h \) and \( w \) denote the height and width of the input image, \( p \) represents the patch size, and \( d \) is the embedding dimension. These tokens are subsequently flattened into a one-dimensional sequence \(\mathbf{T}_{\text{flat}} \in \mathbb{R}^{N \times d}\), where \( N = \frac{h}{p} \times \frac{w}{p} \), and processed by the transformer encoder \( E(\cdot) \) to generate a single-scale representation:
\begin{equation}
\mathbf{Z} = E(\mathbf{X}).
\end{equation}
This representation is then passed through a transformer decoder \( D(\cdot) \), where each token is projected back to pixel space to reconstruct the complete image:
\begin{equation}
\hat{\mathbf{X}} = D(\mathbf{Z}).
\end{equation}
For an autoencoder, the training objective is to minimize the discrepancy between the reconstructed image \(\hat{\mathbf{X}}\) and the original input image \(\mathbf{X}\). This is achieved through a composite loss function:
\begin{equation}
\mathcal{L} = \mathcal{L}_{\text{rec}} + \lambda_1 \mathcal{L}_p + \lambda_2 \mathcal{L}_g + \lambda_3 \mathcal{L}_{\text{kl}},
\label{equation:supervision1}
\end{equation}
where \(\mathcal{L}_{\text{rec}}\) computes the reconstruction loss, typically using a combination of \( L_1 \) loss and MSE loss:
\begin{equation}
\mathcal{L}_{\text{rec}} = \beta_1 \mathcal{L}_1 + \beta_2 \mathcal{L}_2.
\label{equation:supervision2}
\end{equation}
\(\mathcal{L}_p\) represents a perceptual loss~\citep{lpips} captureing high-level perceptual differences, and \(\mathcal{L}_g\) is a discriminative loss, like the discriminator loss in StyleGAN~\citep{gan, stylegan}, which enhances the realism of the generated images. The KL divergence loss~\citep{vae} \(\mathcal{L}_{\text{kl}}\) regularizes the latent distribution to approximate a standard normal distribution.

\paragraph{Limitations of Single-Scale Representation:} 
The vanilla ViT tokenizer operates on a single scale, transforming an image into a fixed grid of tokens, which restricts its capacity to capture hierarchical features. In contrast, convolutional tokenizers naturally encode multi-scale information through a fine-to-coarse compression in the encoder and a coarse-to-fine reconstruction in the decoder, preserving both global context and local details. To address the limitations of the vanilla ViT, we propose a multi-scale design that enhances its capability to represent hierarchical features.

\subsection{Multi-Scale ViT Tokenizer Design}
As shown in Figure \ref{figure:architecture}, we propose a multi-scale ViT tokenizer with minimal architectural modifications, addressing two critical aspects: the construction and the interaction of multi-scale representations. To ensure scale consistency, we employ multi-scale RGB supervision during training.

\paragraph{Multi-Scale Tokens Construction}
To overcome the limitation of fixed token counts in vanilla ViT, we introduce a multi-scale downsampling module \( \mathcal{D}(\cdot) \) following the encoder \( E(\cdot) \). Given the initial single-scale token map \( \mathbf{Z}_0 \in \mathbb{R}^{\frac{H}{p} \times \frac{W}{p} \times d} \) produced by the encoder, we generate a set of token maps \( \{ \mathbf{Z}_s \}_{s=0}^S \) at different scales. Each token map \( \mathbf{Z}_s \in \mathbb{R}^{N_s \times d} \) corresponds to a specific scale \( s \), where \( N_s = \frac{H}{p \cdot 2^{S-s}} \times \frac{W}{p \cdot 2^{S-s}} \). These token maps are then concatenated sequentially from \textbf{low to high resolution} to form a multi-scale representation:
\begin{equation}
\mathbf{Z}_{\text{multi}} = \text{Concat}(\mathbf{Z}_0, \mathbf{Z}_1, \dots, \mathbf{Z}_S) \in \mathbb{R}^{(\sum_{s=0}^S N_s) \times d}.
\end{equation}

For the downsampling module \( \mathcal{D}(\cdot) \), we explore two strategies: (1) \textbf{interpolation-based downsampling}, where \( \mathbf{Z}_s \) is obtained by interpolating \( \mathbf{Z}_S \) using area-pooling methods, and (2) \textbf{learnable convolutional downsampling}, where \( \mathbf{Z}_s \) is generated by applying a trainable convolutional layer \( \text{Conv}_s(\cdot) \) to \( \mathbf{Z}_S \). Both downsampling methods demonstrate performance gains in reconstruction tasks. Nevertheless, for generation tasks, only the convolution-based downsampling approach exhibits the capability to improve generation quality, while interpolation-based downsampling method does not yield such benefits. A detailed ablation study in Section~\ref{section:downsample} further analyzes these trade-offs.

\begin{wrapfigure}{r}{0.4\textwidth}  
    \centering
    \includegraphics[width=0.45\textwidth]{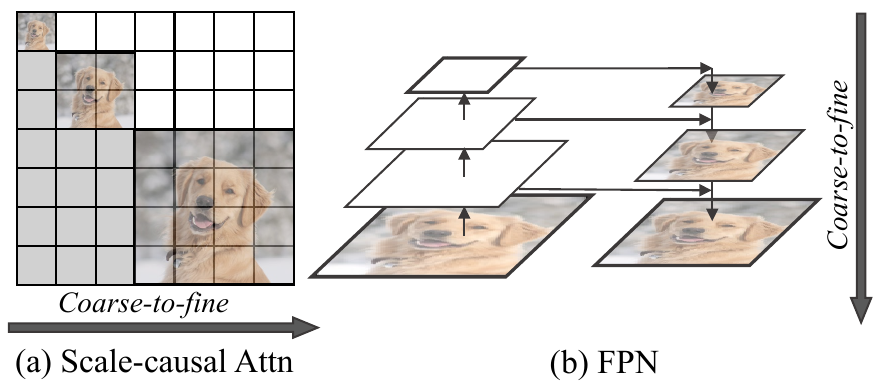}
    \caption{Our scale-causal attention operates akin to FPN along the token sequence dimension.
    }
    \label{fig:attention}
\end{wrapfigure}

\begin{table*}[t]
\centering
\renewcommand{\arraystretch}{0.9}
\resizebox{\textwidth}{!}{
\begin{tabular}{lcccc}
\toprule
\textbf{Tokenizer} & \textbf{Generative Models} & \textbf{rFID $\downarrow$} & \textbf{gFID(w/o CFG) $\downarrow$} & \textbf{gFID(w CFG) $\downarrow$} \\
\midrule
\multicolumn{5}{l}{\textit{Conv-based Tokenizer}} \\
\midrule
LDM\citep{ldm} & MAR\citep{mar}  & 0.53 & 2.35 & 1.55 \\
SD-VAE\citep{sd-vae-ft-ema} & REPA\citep{repa}  & 0.61 & 5.90 & 1.42 \\
SD-VAE & MDT\citep{mdt}  & 0.61 & 6.23 & 1.79 \\
SD-VAE & MDTv2\citep{mdtv2}  & 0.61 & - & 1.58 \\
SD-VAE & DiT\citep{dit}  & 0.61 & 9.62 & 2.27 \\
SD-VAE & SiT\citep{sit}  & 0.61 & 8.61 & 2.06 \\
SD-VAE & MaskDiT\citep{maskdit}  & 0.61 & 5.69 & 2.28 \\
VAR-VAE\citep{var} & VAR-d30\citep{var} & 1.00 & - & 1.92 \\
VA-VAE\citep{vavae} & LightningDiT\citep{vavae}  & 0.28 & 2.17 & 1.35 \\
\midrule
\multicolumn{5}{l}{\textit{ViT-based Tokenizer}} \\
\midrule
MaskGIT\citep{maskgit} & MaskGIT\citep{maskgit}  & 2.28 & 6.18 & - \\
VQGAN\citep{vqgan} & LlamaGen\citep{llamagen} & 0.59 & 9.38 & 2.18 \\
FlexTok d18-d28\citep{flextok} & LlamaGen\citep{llamagen}  & 1.45 & - & 1.86 \\
ViTok S-L\citep{scaling_vit} & LlamaGen\citep{llamagen}  & 0.46 & - & 2.45 \\
TiTok-S\citep{titok} & MaskGIT\citep{maskgit}  & 1.71 & - & 1.97 \\
MAETok\citep{maetok} & LightningDiT\citep{vavae} & 0.48 & - & 1.73 \\
GigaTok-XL-XXL\citep{gigatok} & LlamaGen-XXL\citep{llamagen} & 0.79 &  - & 1.98 \\
HieraTok & DiT  & 0.45 & 3.53 & 1.82 \\
\bottomrule
\end{tabular}
}
\caption{Performance of HieraTok in the Context of sota Generative Models on ImageNet 256x256. HieraTok achieves a competitive rFID of 0.45 and gFID of 1.82 for ViT-based tokenizers.}
\label{tab:scaleup results}
\end{table*}

\paragraph{Multi-Scale Information Interaction} In the decoder, we design two attention mechanisms to facilitate interaction among multi-scale token representations. The first, termed \textbf{scale-independent attention}, restricts each token to interact only with tokens within the same scale, ensuring independence between representations at different resolutions. Mathematically, this can be expressed as:
\begin{equation}
\text{Attention}(\mathbf{Q}_s, \mathbf{K}_s, \mathbf{V}_s) = \text{Softmax}\left(\frac{\mathbf{Q}_s \mathbf{K}_s^\top}{\sqrt{d}}\right) \mathbf{V}_s,
\end{equation}
where \( \mathbf{Q}_s \), \( \mathbf{K}_s \), and \( \mathbf{V}_s \) denote the query, key, and value matrices for the \( s \)-th scale, respectively.
The second mechanism, \textbf{scale-causal attention}, allows tokens at the current resolution to interact with tokens at both the current and preceding resolutions. Combined with our token concatenation strategy, which proceeds from low to high resolution, this design enables a hierarchical information flow, where high-level semantic features from lower resolutions progressively refine low-level structural details at higher resolutions. Mathematically, this is formulated as:
\begin{equation}
\text{Attention}(\mathbf{Q}_s, \mathbf{K}_{\leq s}, \mathbf{V}_{\leq s}) = \text{Softmax}\left(\frac{\mathbf{Q}_s \mathbf{K}_{\leq s}^\top}{\sqrt{d}}\right) \mathbf{V}_{\leq s},
\end{equation}
where \( \mathbf{K}_{\leq s} \) and \( \mathbf{V}_{\leq s} \) represent the key and value matrices for all resolutions up to and including the \( s \)-th scale.
This information propagation mechanism is analogous to the top-down feature aggregation in Feature Pyramid Networks (FPNs) as shown in Figure \ref{fig:attention}. By integrating this multi-scale information flow into the transformer's token interactions, each attention operation simultaneously performs a function similar to FPN's feature aggregation along the token sequence dimension.

\paragraph{Multi-Scale RGB Supervision}
During decoding, multi-scale tokens, like their single-scale counterparts, are regressed to reconstruct the RGB pixels of their corresponding patches. Specifically, we downsample the ground truth image $ \mathbf{X} $ to match each scale $ s $ using an interpolation operation $ \mathcal{D}_s(\cdot) $, and supervise the reconstruction of the corresponding resolution. This supervision effectively introduces \textbf{scale equivariance} into the representation space and the decoder. Formally, let $ \mathcal{D}_s(\cdot) $ denote the downsampling operation for scale $ s $. We establish the following optimization objectives for reconstruction:
\begin{equation}
    \mathcal{L}_{\text{full}} = \| D(\mathbf{Z}) - \mathbf{X} \|_2^2, \quad \text{where} \quad \mathbf{Z} = E(\mathbf{X}),
\end{equation}
\begin{equation}
    \mathcal{L}_{\text{scale}} = \| D(\mathcal{D}_s(\mathbf{Z})) - \mathcal{D}_s(\mathbf{X}) \|_2^2.
\end{equation}
where $ E(\cdot) $ is the encoder and $ D(\cdot) $ is the decoder. This supervision ensures that downsampling operations commute with the latent representation and decoder (i.e., $ D(\mathcal{D}_s(\mathbf{Z})) \approx \mathcal{D}_s(D(\mathbf{Z})) $), thereby enforcing scale consistency across resolutions.

We validate that the multi-scale supervised tokenizer achieves significant improvements in both reconstruction and generation performance, while the introduced scale consistency likely leads to better latent space structure and more uniform distributions.

\begin{table*}[t]
\centering
\renewcommand{\arraystretch}{1.1}
\resizebox{\textwidth}{!}{
\begin{tabular}{lcccccc|cc}
\toprule
\multirow{2}{*}{\textbf{Spec.}} & \multirow{2}{*}{\textbf{KL Weight}} & \multirow{2}{*}{\textbf{Tokenizer}} & \multicolumn{4}{c|}{\textbf{Reconstruction}} & \multicolumn{2}{c}{\textbf{Generation}} \\
\cmidrule(lr){4-7} \cmidrule(lr){8-9}
& & & \textbf{rFID}$\downarrow$ & \textbf{PSNR$\uparrow$} & \textbf{LPIPS$\downarrow$} & \textbf{SSIM$\uparrow$} & \textbf{gFID$\downarrow$} & \textbf{IS$\uparrow$} \\
\midrule
\multirow{2}{*}{f16d16} & \multirow{2}{*}{1e-6} & Single Scale & \cellcolor{blue!20}1.47 & 24.4 & 0.108 & 0.74 & \multicolumn{1}{c}{\cellcolor{blue!20}16.4} & 57.4 \\
 & & Multi Scale & 1.07\textcolor{orange}{\textit{(-0.40)}} & 23.8 & 0.087 & 0.72 & 13.3 \textcolor{orange}{\textit{(-3.1)}} & 68.5 \textcolor{gray}{\textit{(+11.1)}} \\
 \midrule
 \multirow{2}{*}{f16d16} & \multirow{2}{*}{1e-5} & Single Scale & \cellcolor{blue!20}1.76 & 23.9 & 0.122 & 0.72 & \multicolumn{1}{c}{\cellcolor{blue!20}20.2} & 51.0 \\
 & & Multi Scale & 1.61\textcolor{orange}{\textit{(-0.16)}} & 22.8 & 0.115 & 0.67 & 17.9 \textcolor{orange}{\textit{(-2.3)}} & 55.5 \textcolor{gray}{\textit{(+4.5)}} \\
  \midrule
 \multirow{2}{*}{f16d32} & \multirow{2}{*}{1e-5} & Single Scale & \cellcolor{blue!20}1.60 & 24.3 & 0.119 & 0.73 & \multicolumn{1}{c}{\cellcolor{blue!20}24.3} & 43.6 \\
 & & Multi Scale & 1.54\textcolor{orange}{\textit{(-0.06)}} & 23.0 & 0.113 & 0.68 & 19.4 \textcolor{orange}{\textit{(-4.9)}} & 52.5 \textcolor{gray}{\textit{(+8.9)}} \\
\bottomrule
\end{tabular}}
\caption{Comparative evaluation of multi-scale versus single-scale tokenizers under various settings reveals the multi-scale variant's consistent performance advantages in both image reconstruction and generation tasks. We highlight the baseline's rFID and gFID in blue for better distinction.}
\label{tab:main results}
\end{table*}

\begin{figure}[t]
  \centering
  \includegraphics[width=1.0\linewidth]{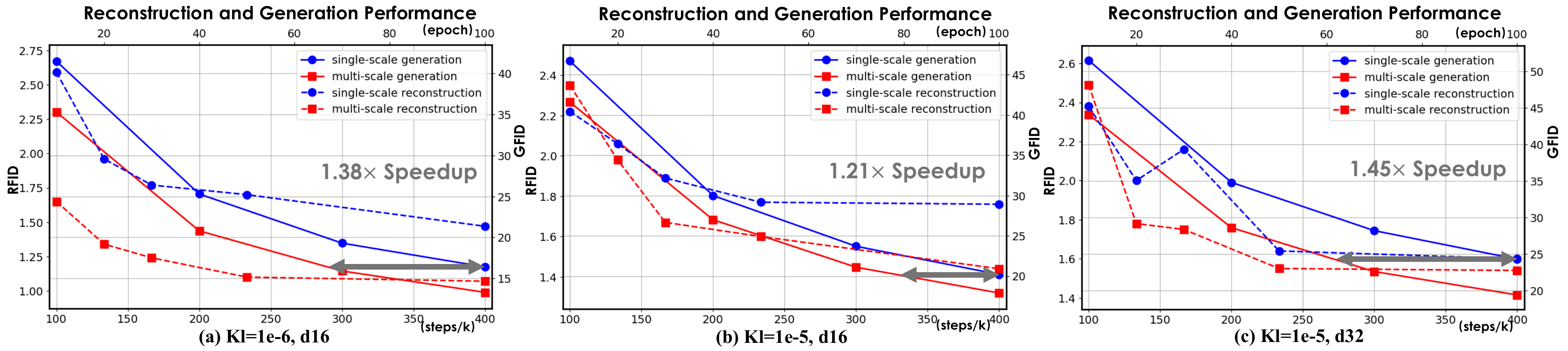}
  \caption{Performance evolution of multi-scale versus single-scale tokenizers across training iterations for both reconstruction and generation tasks. The multi-scale tokenizer demonstrates consistent superiority over its single-scale counterpart throughout the training process.}
  \label{figure:main results}
\end{figure}

\begin{table*}[t]
\centering
\renewcommand{\arraystretch}{1.2}
\resizebox{\textwidth}{!}{
\begin{tabular}{lllcccc|cc}
\toprule
\multirow{2}{*}{\textbf{Tokenizer}} & \multirow{2}{*}{\textbf{Downsample}} & \multirow{2}{*}{\textbf{Attention}} & \multicolumn{4}{c|}{\textbf{Reconstruction}} & \multicolumn{2}{c}{\textbf{Generation}} \\
\cmidrule(r){4-7} \cmidrule(l){8-9}
 & & & \textbf{rFID}$\downarrow$ & \textbf{PSNR}$\uparrow$ & \textbf{LPIPS}$\downarrow$ & \textbf{SSIM}$\uparrow$ & \textbf{gFID}$\downarrow$ & \textbf{IS}$\uparrow$ \\
\midrule

Single-Scale & - & - & \cellcolor{blue!20}1.47 & 24.4 & 0.108 & 0.74 & \cellcolor{blue!20}16.4 & 57.4 \\

\midrule

\multirow{6}{*}{Multi-Scale}
& Conv & Scale-causal & \textbf{1.07} & 23.8 & 0.087 & 0.72 & \textbf{13.3} & 68.5 \\
& Conv & Scale-independent & 1.13 & 23.7 & 0.093 & 0.71 & 14.0 & 65.3 \\
& Conv & Full-Attn & 1.24 & 23.6 & 0.099 & 0.71 & 16.6 & 56.4 \\
& Interpolate & Scale-causal & 1.18 & 23.7 & 0.095 & 0.71 & 17.8 & 52.1 \\
& Interpolate & Scale-independent & 1.16 & 23.7 & 0.091 & 0.71 & 17.2 & 54.4 \\
& Interpolate & Full-Attn & 1.13 & 23.0 & 0.113 & 0.68 & 16.9 & 55.7 \\

\bottomrule
\end{tabular}}
\caption{Performance comparison of multi-scale tokenizer variants with different downsampling and attention designs.}
\label{tab:downsample&attention}
\end{table*}

\begin{table*}[t]
\centering
\renewcommand{\arraystretch}{1.0}
\resizebox{0.9\textwidth}{!}{
\begin{tabular}{llcccc|cc}
\toprule
\multirow{2}{*}{\textbf{Tokenizer}} & \multirow{2}{*}{\textbf{Scales}} & \multicolumn{4}{c|}{\textbf{Reconstruction}} & \multicolumn{2}{c}{\textbf{Generation}} \\
\cmidrule(lr){3-6} \cmidrule(lr){7-8}
 & & \textbf{rFID}$\downarrow$ & \textbf{PSNR}$\uparrow$ & \textbf{LPIPS}$\downarrow$ & \textbf{SSIM}$\uparrow$ & \textbf{gFID}$\downarrow$ & \textbf{IS}$\uparrow$ \\
\midrule

Single-Scale & [16] & \cellcolor{blue!20}1.47 & 24.4 & 0.108 & 0.74 & \cellcolor{blue!20}16.4 & 57.4 \\

\midrule

\multirow{3}{*}{Multi-Scale}
& [1,2,4,16] & 1.25 &  23.6 & 0.092 & 0.71 & 15.9 & 60.2 \\
& [1,2,4,8,16] & 1.07 & 23.8 & 0.087 & 0.72 & \textbf{13.3} & 68.5 \\
& [1,2,4,8,12,16] & \textbf{1.04} & 23.9 & 0.087 & 0.72 & 13.8 & 66.8 \\

\bottomrule
\end{tabular}}
\caption{Performance comparison of multi-scale tokenizer variants with different multi-scale setting.}
\label{tab:scale_analysis}
\end{table*}

\begin{figure}[t]
  \centering
  \includegraphics[width=1.0\linewidth]{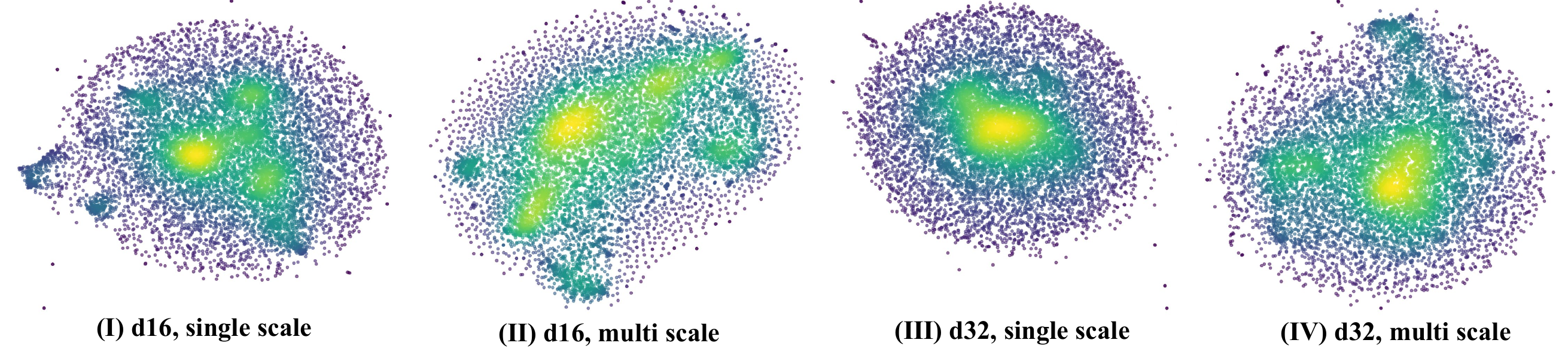}
  \caption{\textbf{t-SNE visualization of latent space representations}, demonstrating that the multi-scale tokenizer achieves more uniform feature distribution compared to the single-scale baseline.}
  \label{figure:tsne}
\end{figure}

\begin{table*}[t]
\centering
\renewcommand{\arraystretch}{1.0}
\resizebox{0.9\textwidth}{!}{
\begin{tabular}{ll|cccc}
\toprule
\textbf{Spec.} & \textbf{Tokenizer} & \textbf{Density CV}$\downarrow$ & \textbf{Gini Coefficient}$\downarrow$ & \textbf{Normalized Entropy}$\uparrow$ & \textbf{gFID}$\downarrow$ \\
\midrule

\multirow{2}{*}{f16d16} & Single-Scale & 0.522 & 0.299 & 0.984 & 16.4 \\
 & Multi-Scale & 0.381 & 0.216 & 0.991 & 13.3 \\
\midrule
\multirow{2}{*}{f16d32} & Single-Scale & 0.656 & 0.371 & 0.976 & 24.3 \\
 & Multi-Scale & 0.435 & 0.249 & 0.988 & 19.4 \\

\bottomrule
\end{tabular}}
\caption{\textbf{Analysis of latent space uniformity}, demonstrating that the multi-scale design achieves more uniform latent distributions while simultaneously improving generation quality.}
\label{tab:tsne}
\end{table*}

\section{Implementation Details}
\paragraph{Tokenizer Training} Following the conclusions from ~\citep{scaling_vit}, we adopt a small ViT encoder and a large ViT decoder for all tokenizer architectures. All tokenizers are trained on the ImageNet 256$\times$256 dataset~\citep{imagenet} with a learning rate of\( 1 \times 10^{-4} \) and a global batch size of 256. We evaluate two compression ratios: \textbf{f16d16} and \textbf{f16d32}, where \textbf{f} represents the spatial compression ratio and \textbf{d} denotes the dimension of latent channels. For the multi-scale tokenizer, the default multi-scale design is implemented with scales [1,2,4,8,16]. For our ablation studies, we train models for 100 epochs using only L1, L2, perceptual, and KL losses to obtain results quickly. For the scaled-up tokenizer experiments, we supplement these with GAN loss and vf-loss and increase the number of training epochs to achieve better performance.

\paragraph{Performance Evaluation} We assess the reconstruction quality of the tokenizer using metrics including FID~\citep{fid}, PSNR~\citep{psnr}, and SSIM~\citep{ssim} on the ImageNet-50k validation dataset. To validate the performance of different tokenizers on generative tasks, we conduct extensive experiments using the \textbf{vanilla DiT} framework as our testbed. Specifically, we employ DiT-XL as our default generative model with a global batch size of 256 and a constant learning rate of 1e-4 without decay. For generative performance evaluation, we generate 50k images without classifier-free guidance (CFG)~\citep{cfg} and compute both gFID and IS~\citep{IS} metrics. For our ablation studies, all tokenizer variants are trained on DiT-XL~\citep{dit} for \textbf{400k steps}. For the scaled-up experiments, we train for \textbf{5M} steps to achieve superior generation performance.

\begin{table*}[t]
\centering
\renewcommand{\arraystretch}{1.0}
\resizebox{0.8\textwidth}{!}{
\begin{tabular}{cccc|cc}
\toprule
\multirow{2}{*}{\textbf{Generation Model}} & \multirow{2}{*}{\textbf{Type}} & \multirow{2}{*}{\textbf{Params}} & \multirow{2}{*}{\textbf{Tokenizer}} & \multicolumn{2}{c}{\textbf{Generation Performance}} \\
\cmidrule(lr){5-6}
& & & & \textbf{gFID$\downarrow$} & \textbf{IS$\uparrow$} \\
\midrule
\multirow{2}{*}{DiT-XL} & \multirow{2}{*}{Diff} & \multirow{2}{*}{675M} & Single Scale & \multicolumn{1}{c}{\cellcolor{blue!20}16.4} & 57.4 \\
 & & & Multi Scale & 13.3 \textcolor{orange}{\textit{(-3.1)}} & 68.5 \textcolor{gray}{\textit{(+11.1)}} \\
\midrule
\multirow{2}{*}{DiT-L} & \multirow{2}{*}{Diff} & \multirow{2}{*}{458M} & Single Scale & \multicolumn{1}{c}{\cellcolor{blue!20}19.3} & 50.7 \\
 & & & Multi Scale & 16.5 \textcolor{orange}{\textit{(-2.8)}} & 58.9 \textcolor{gray}{\textit{(+8.2)}} \\
\midrule
\multirow{2}{*}{DiT-B} & \multirow{2}{*}{Diff} & \multirow{2}{*}{130M} & Single Scale & \multicolumn{1}{c}{\cellcolor{blue!20}33.6} & 31.8 \\
 & & & Multi Scale & 30.8 \textcolor{orange}{\textit{(-2.8)}} & 35.2 \textcolor{gray}{\textit{(+3.4)}} \\
\bottomrule
\end{tabular}}
\caption{\textbf{Generation performance evaluation across different DiT model sizes}, demonstrating consistent superiority of the multi-scale tokenizer over the single-scale baseline.}
\label{tab:genmodel size}
\end{table*}

\section{Experiments Results and Analysis}
\subsection{Multi-Scale Tokenizer Improves Generation and Reconstruction}

Our results demonstrate that the multi-scale tokenizer consistently improves both reconstruction and generation performance over its single-scale counterpart across different KL divergence weights (1e-5 and 1e-6) and latent compression ratios (f16d16 and f16d32) settings as shown in Figure \ref{figure:main results} and Table \ref{tab:main results}. For reconstruction, under the KL=1e-6, d16 compression setting, HieraTok achieves a 27.2\% reduction in rFID and a 19.4\% improvement in LPIPS (Table 2), highlighting its superior perceptual fidelity. For generation, under strictly controlled training settings (400k steps), the multi-scale tokenizer not only converges 1.38x faster but also yields a significant 18.9\% gFID improvement ($16.4 \rightarrow 13.3$), as shown in Figure \ref{figure:main results}. We further validate this advantage across various DiT model sizes (Table \ref{tab:genmodel size}), confirming the robustness of our multi-scale design.

To situate HieraTok within the state-of-the-art landscape, we also conduct a scaled-up experiment using an enhanced training strategy over 5M steps. The resulting model achieves a new sota rFID of 0.45 among ViT-based tokenizers and a highly competitive gFID of 1.82 (Table \ref{tab:scaleup results}). This result underscores the high performance ceiling of our proposed architecture, complementing the direct, controlled comparisons that validate the effectiveness of the core design.

\subsection{Visualization and Analysis of Latent Space Structure}
To investigate the intrinsic advantages of our multi-scale design, we conduct a comprehensive latent space analysis using t-SNE~\citep{tsne} visualization in Figure \ref{figure:tsne}. We analyze the latent distributions of 10k ImageNet test images using t-SNE projections in Table \ref{tab:tsne}, following VAVAE's~\citep{vavae} methodology by calculating the standard deviation and Gini coefficients through kernel density estimation. Both qualitative visualizations and quantitative metrics demonstrate that the multi-scale tokenizer achieves more uniform latent space distributions. Notably, we observe that increasing the tokenizer's latent dimension leads to over-concentrated clusters, potentially explaining the training difficulties in high-dimensional settings, while our multi-scale architecture consistently alleviates this issue. These findings collectively suggest that the multi-scale tokenizer's superior performance, particularly its faster convergence in generation tasks, stems from its better-structured latent geometry and more homogeneous distribution properties.

\subsection{Analysis on Multi-Scale Downsample Methods}
\label{section:downsample}
As shown in Table \ref{tab:downsample&attention}, we investigated two distinct downsampling strategies: a parameter-free approach using direct interpolation and a learnable approach based on a series of convolutional kernels. Both strategies led to substantial gains in reconstruction fidelity. However, we observed a critical performance trade-off when applying these models to generative tasks. The learnable convolutional method consistently improved generation quality, while the direct interpolation strategy's performance was not only significantly inferior but also failed to surpass the single-scale generation baseline.

We hypothesize that the divergent outcomes stem from the fundamental differences between the two tasks. In reconstruction, the model benefits directly from multi-scale supervision, where each level of the decoder is guided by a corresponding ground-truth image scale. This multi-level guidance is the primary source of improvement, placing less stringent demands on the latent representation itself. Generation, however, is critically dependent on the learned latent distribution. We postulate that a simple interpolation-based downsampler cannot effectively differentiate between high-level semantic content and low-level visual textures. This makes it difficult to form a well-structured, hierarchical representation with distinct levels of feature granularity. Therefore, learnable convolutional kernels are essential, as they actively learn transformations that separate and specialize the visual information appropriate for each scale.

\subsection{Analysis on Multi-Scale Attention Mechanism}

We conducted an ablation study on three different attention mechanisms under the KL=1e-6, d16 setting: (1) the standard full-attention from the original ViT; (2) our proposed scale-causal attention; and (3) a scale-independent variant. The results in Table \ref{tab:downsample&attention} show that while all three mechanisms improve reconstruction quality, their generative performance varies considerably. The scale-causal attention markedly outperforms both the full-attention and scale-independent mechanisms in generation tasks. 

We attribute the reconstruction improvements to the inherent benefits of multi-scale representations and supervision, as discussed in the previous Section \ref{section:downsample}. Generative modeling, however, imposes stricter requirements on the structure of the latent representations. Although full-attention allows for bidirectional information flow between coarse and fine scales, its unrestricted visibility makes it difficult to differentiate between scale-specific features. This limits its ability to effectively model the coarse-to-fine progression from global semantics to local details, resulting in generative performance comparable to a single-scale tokenizer. Conversely, the scale-independent mechanism successfully isolates representations at each scale but lacks the inter-scale interaction needed to fully leverage the multi-scale advantage. Only the scale-causal design, which models a progressive refinement process analogous to FPN where low-resolution semantic features inform high-resolution structural features, effectively optimizes the latent space structure to yield substantial gains in generation quality.

\subsection{Analysis on Multi-Scale Settings}
Through systematic investigation of multi-scale configurations under the kl=1e-6 and f16d16 parameter settings, we evaluate three distinct scale combinations and assess their impact on both reconstruction and generation performance in Table \ref{tab:scale_analysis}. Our experiments demonstrate that incorporating just three scales (1, 2, 4), corresponding to 21 tokens, yields a significant 15\% improvement in reconstruction quality (from 1.47 to 1.25), with further scale augmentation consistently enhancing reconstruction performance. For generation tasks, while multi-scaling improves results, we observe diminishing returns beyond five scales (1, 2, 4, 8, 16), suggesting this combination may represent the performance ceiling for this approach. Based on these empirical findings, we recommend the (1, 2, 4, 8, 16) configuration as optimal for balancing computational efficiency with performance gains.

\section{Conclusion}
In this work, we introduce a novel multi-scale ViT tokenizer HieraTok that overcomes the limitation of vanilla ViT in modeling only single-scale token maps. Through meticulous design of multi-scale tokens construction and inter-scale information interaction mechanisms, we demonstrate that our multi-scale tokenizer significantly outperforms its single-scale counterpart in both reconstruction and generation tasks. Furthermore, our investigation reveals that multi-scale representation and supervision introduce scale consistency, which regularizes the latent space structure, resulting in a smoother and more uniform latent space distribution. This property may explain why the multi-scale design accelerates convergence in generation tasks. Collectively, these findings strongly suggest that our proposed multi-scale ViT tokenizer represents a substantial advancement over existing ViT-based tokenizer architectures. We believe this method holds great promise as a standard strategy for training high-performance tokenizers for image reconstruction and generation.

\section{Acknowledgment}
This work was supported by Ant Group Research Intern Program.

\bibliography{main}
\bibliographystyle{main}

\clearpage
\appendix
\section{Use of LLM}
\textbf{The use of a Large Language Model (LLM) is limited to language editing and refinement.}

\section{Derivation of the Decoder-Downsampling Approximation}
\label{sec:appendix_approximation}

This section provides a detailed derivation for the approximation $D(D_s(Z)) \approx D_s(D(Z))$. This relationship, which suggests that the decoder and downsampling operations are approximately commutative, is foundational to our multi-scale supervision strategy. The validity of this approximation is contingent on a well-trained tokenizer capable of high-fidelity image reconstruction.

Let $X$ represent a high-resolution image, and let $Z = E(X)$ be its corresponding latent representation obtained from the encoder $E(\cdot)$. Let $D(\cdot)$ be the decoder, and let $D_s(\cdot)$ be the downsampling operation for a given scale $s$. The derivation proceeds as follows:

\begin{enumerate}
    \item \textbf{Multi-Scale Reconstruction Objective:} The multi-scale supervision loss, $\mathcal{L}_{\text{scale}}$, is designed to train the decoder to reconstruct a downsampled image, $D_s(X)$, from a correspondingly downsampled latent code, $D_s(Z)$. For an optimally trained model, this objective leads to the following approximation:
    \begin{equation}
        D(D_s(Z)) \approx D_s(X)
        \label{eq:approx_1}
    \end{equation}

    \item \textbf{Full-Resolution Reconstruction Objective:} Simultaneously, the primary reconstruction objective trains the decoder to reconstruct the original full-resolution image $X$ from the complete latent code $Z$. For a well-trained model, this yields:
    \begin{equation}
        D(Z) \approx X
        \label{eq:approx_2}
    \end{equation}

    \item \textbf{Downsampling the Full Reconstruction:} By applying the downsampling operator $D_s(\cdot)$ to both sides of the approximation in Equation~\eqref{eq:approx_2}, we obtain:
    \begin{equation}
        D_s(D(Z)) \approx D_s(X)
        \label{eq:approx_3}
    \end{equation}

    \item \textbf{Establishing Equivalence:} The expressions in Equation~\eqref{eq:approx_1} and Equation~\eqref{eq:approx_3} are both approximately equal to the same term, $D_s(X)$. By equating their left-hand sides, we arrive at the final derived relationship:
    \begin{equation}
        D(D_s(Z)) \approx D_s(D(Z))
    \end{equation}
\end{enumerate}

This derivation formally demonstrates that if a decoder can accurately reconstruct images at both full and reduced scales from their respective latent codes, the operations of decoding and downsampling become interchangeable in practice.

\section{Analysis of Computational Cost and Training Fairness}
\label{sec:appendix_cost}

A detailed comparison of the computational costs and the fairness of the training protocol is provided to ensure full transparency. A distinction is made between the two primary training stages: tokenizer training (reconstruction) and downstream model training (generation).

\subsection{Computational Cost}

Table~\ref{tab:comp_cost} presents a comparison of the key computational metrics for the single-scale and multi-scale tokenizers.

\begin{table}[h!]
\centering
\begin{tabular}{lcccc}
\toprule
\textbf{Tokenizer} & \textbf{Params} & \textbf{GFLOPS} & \textbf{Encoder Tokens} & \textbf{Decoder Tokens} \\
\midrule
single scale & 346.25M & 182.36 & 256 & 256 \\
multi scale  & 401.18M & 230.45 & 256 & 341 \\
\bottomrule
\end{tabular}
\caption{Comparison of computational cost for single-scale and multi-scale tokenizers during the reconstruction training phase.}
\label{tab:comp_cost}
\end{table}

During the \textbf{tokenizer training (reconstruction) stage}, the multi-scale model exhibits a higher computational cost. This is primarily attributed to its decoder, which processes a longer sequence of concatenated tokens from different scales.

However, for the \textbf{downstream model training (generation) stage}, the methodology is designed to introduce \textbf{zero additional computational overhead}. For the generation task, the latent representation is extracted from the encoder \textit{before} the multi-scale downsampling is applied. This ensures that the latent code passed to the subsequent generative model (e.g., a DiT) has the exact same dimensions as the code from the single-scale baseline. Consequently, both the training and inference processes of the downstream generation task remain unaffected in terms of computational cost and token count.

\subsection{Training Fairness}

To ensure a fair comparison, the training setups for both tokenizer architectures were closely aligned. Both the single-scale and multi-scale tokenizers utilize identical encoder architectures. Furthermore, both encoders process the same set of images at the same resolution. By training both models for the same number of epochs, we guarantee that each architecture is exposed to the exact same volume and nature of primary visual data from the encoder's perspective. Given this controlled setup, comparing the models based on training epochs is a fair and standard method to evaluate the effectiveness of each tokenizer's architecture in learning to reconstruct and represent visual information.

\section{Scale-consistent Generation}
Our HieraTok is trained with multi-scale RGB images as supervision signals, enabling it to generate multi-scale RGB outputs. While the main text only visualizes the largest-scale generated images, we provide additional visualizations in the appendix—showing multi-scale generations using both DiT-XL as the generative model and HieraTok's multi-scale decoder. The consistent generation quality across scales demonstrates that our proposed HieraTok achieves strong scale coherence.

\begin{figure}[h]
  \centering
  \includegraphics[width=1.0\linewidth]{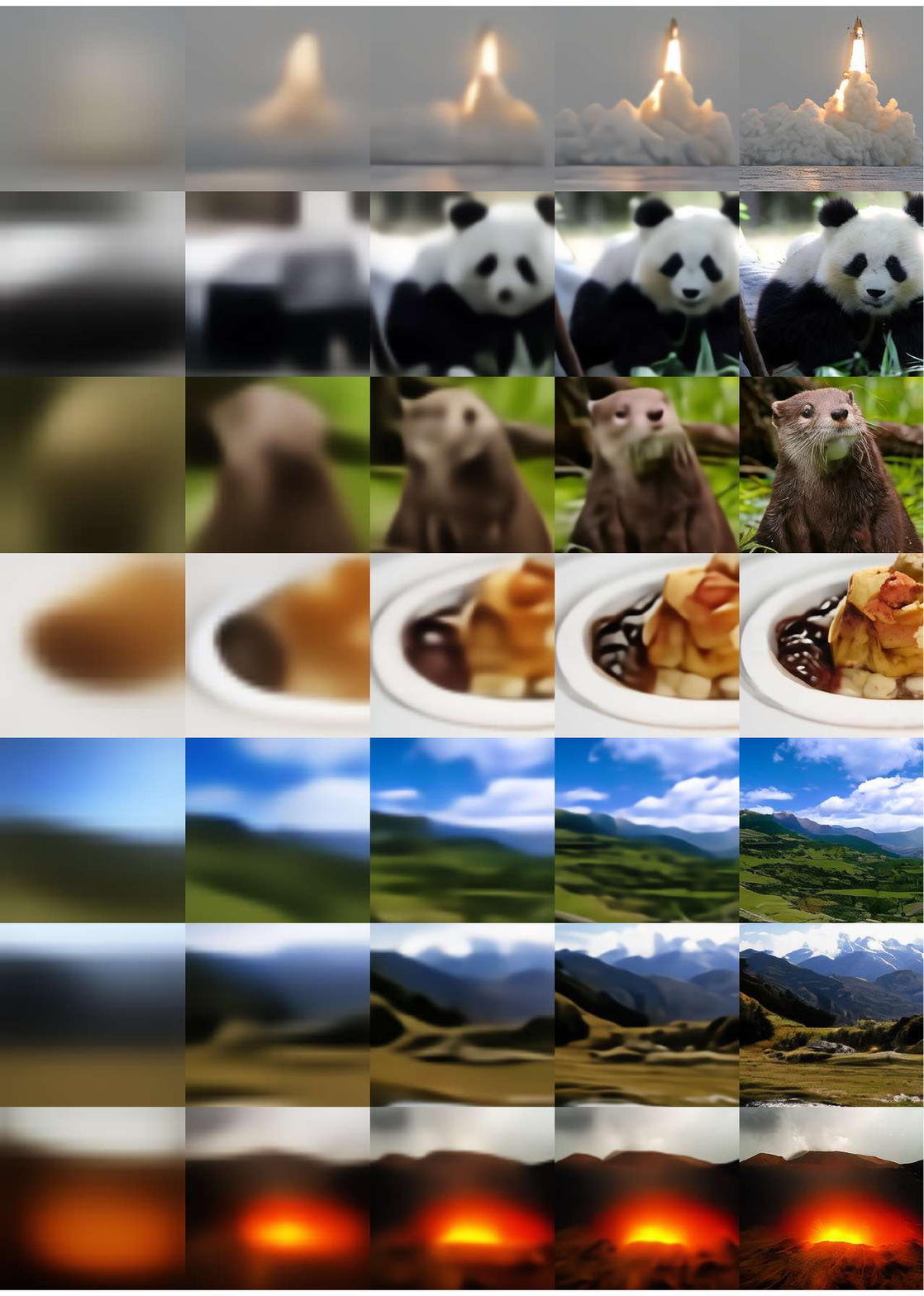}
  \caption{Multi-scale generated samples. Class label 812, 388, 360, 928, 979 and 980.}
\end{figure}

\begin{figure}[h]
  \centering
  \includegraphics[width=1.0\linewidth]{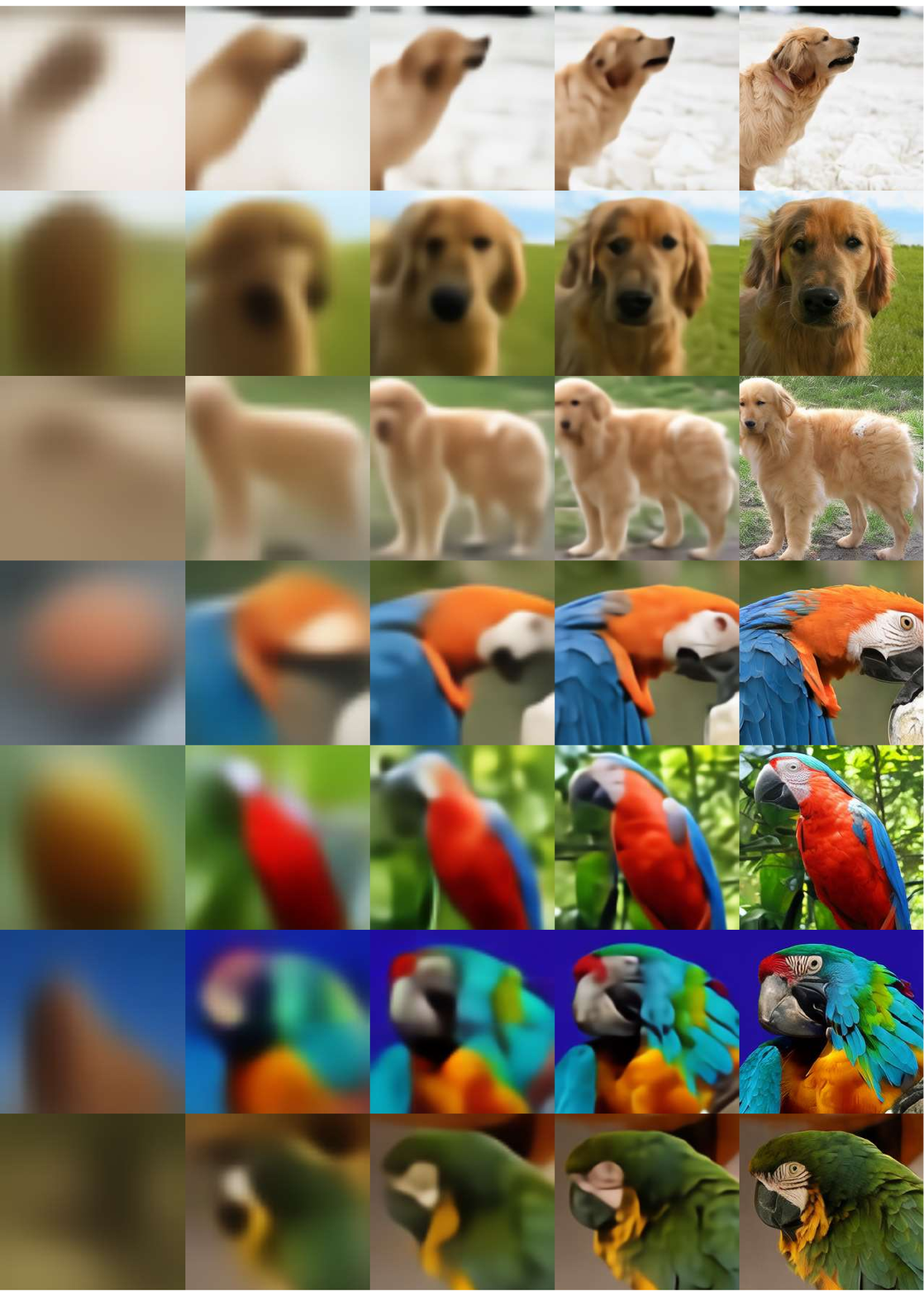}
  \caption{Multi-scale generated samples. Class label 207 and 88.}
\end{figure}

\begin{figure}[h]
  \centering
  \includegraphics[width=1.0\linewidth]{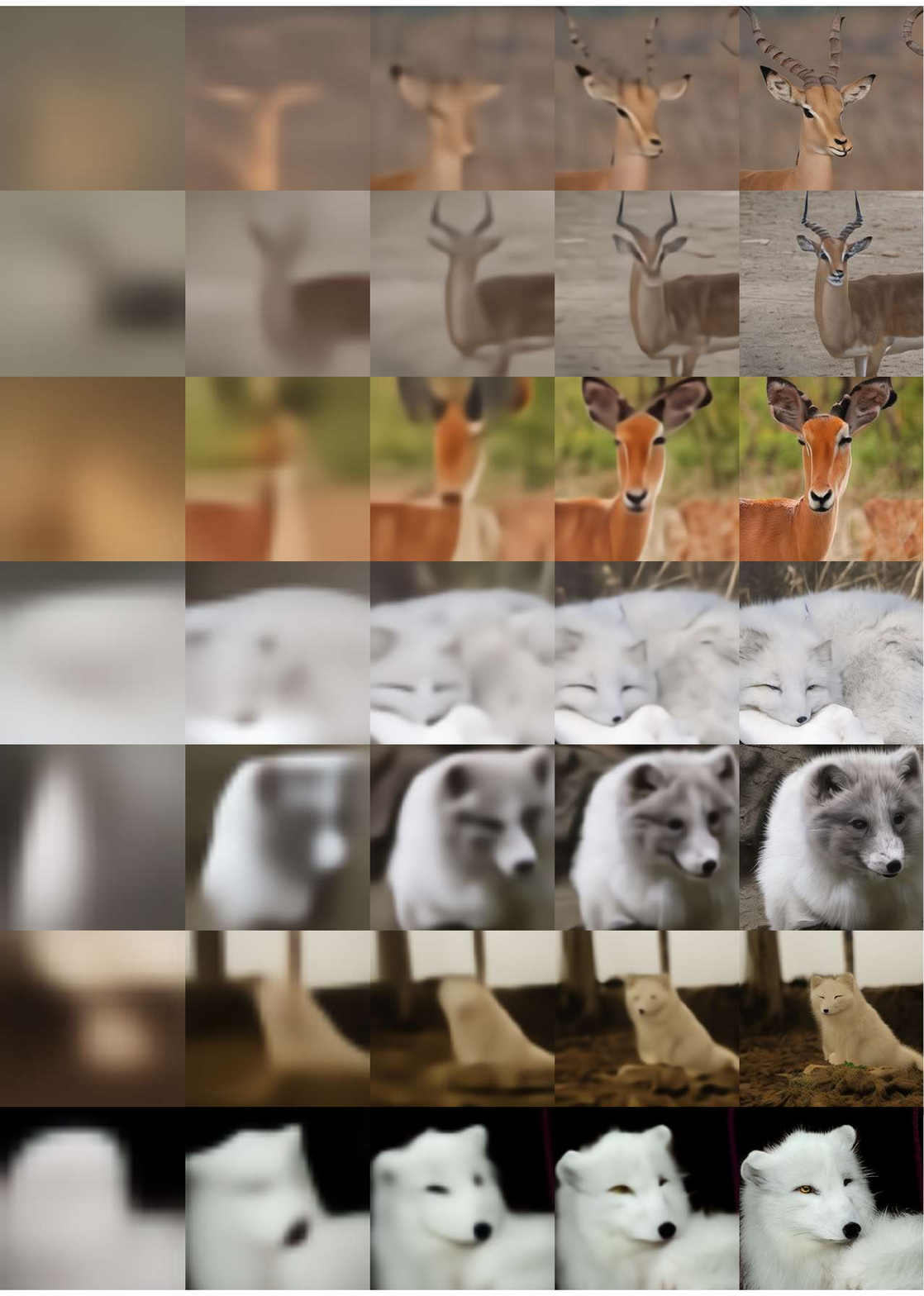}
  \caption{Multi-scale generated samples. Class label 352 and 279.}
\end{figure}

\begin{figure}[h]
  \centering
  \includegraphics[width=1.0\linewidth]{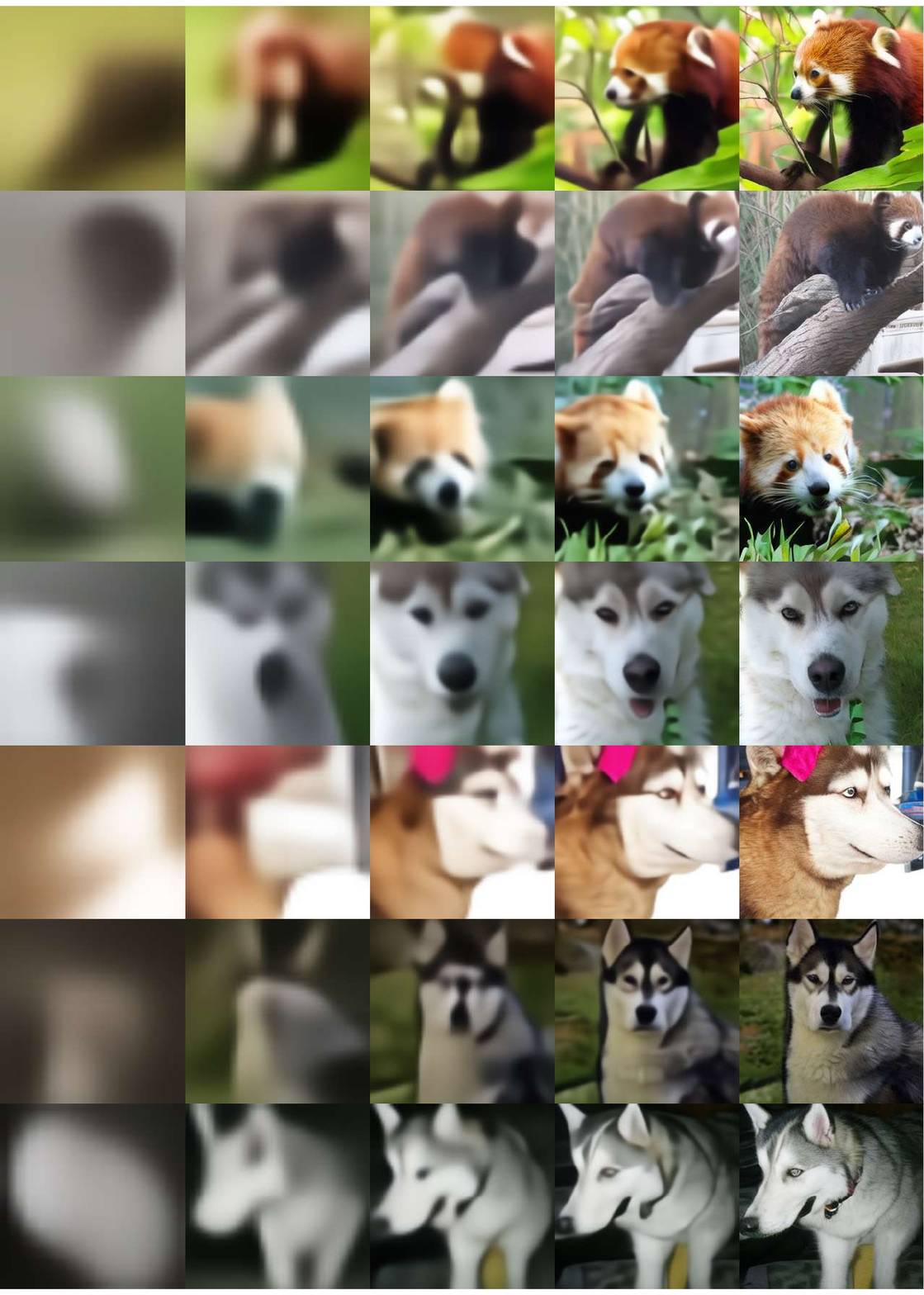}
  \caption{Multi-scale generated samples. Class label 387 and 250.}
\end{figure}

\clearpage
\section{Tokenizer Implementation Details}
\label{sec:tokenizer training}

\begin{table}[h]
    \centering
    \normalsize
    \renewcommand{\arraystretch}{1.2} 
    \begin{tabular}{l|cc}
        \toprule
        \textbf{Parameter} & \textbf{Stage1} & \textbf{Stage2} \\
        \midrule
        Traing data & ImageNet & ImageNet \\
        Training epochs & 100 & 40 \\
        Global Batch size & 256 & 256 \\
        Warmup ratio & 0.03 & 0.03 \\
        Optimizer & AdamW & AdamW  \\
        Learning Rate start & $4.0 \times 10^{-4}$ & $2.0 \times 10^{-4}$  \\
        Learning Rate end & $4.0 \times 10^{-6}$ & $2.0 \times 10^{-6}$  \\
        LR scheduler & CosineAnnealing & CosineAnnealing \\
        Optimizer beta1 & 0.9 & 0.9 \\
        Optimizer beta2 & 0.95 & 0.95 \\
        Weight decay & 0.05  & 0.05 \\
        GAN Optimizer & - & AdamW \\
        GAN LR start & - & $2.0 \times 10^{-4}$ \\
        GAN LR end & - & $2.0 \times 10^{-6}$ \\
        GAN LR scheduler & - & CosineAnnealing  \\
        GAN Warmup ratio & - & 0.01 \\
        GAN Optimizer beta1 & - & 0.9 \\
        GAN Optimizer beta2 & - & 0.95 \\
        GAN Weight decay & -  & $1.0 \times 10^{-4}$ \\
        \bottomrule
    \end{tabular}
    \caption{Training settings for Tokenizer.}
    \label{tab: config of tokenizer training}
\end{table}
\begin{table}[h]
    \centering
    \normalsize
    \renewcommand{\arraystretch}{1.2} 
    \begin{tabular}{l|cc}
        \toprule
        \textbf{Loss Type} & \textbf{Stage1} & \textbf{Stage1} \\
        \midrule
        L1 Loss & 1.0 & 1.0  \\
        MSE Loss & 0.4 & 0.4  \\
        LPIPS Loss & 1.0 & 1.0  \\
        KL Loss & 1e-6 & 1e-6  \\
        GAN Loss & - & 0.6 \\
        \bottomrule
    \end{tabular}
    \caption{Different Loss Weight.}
    \label{tab: config of loss weight}
\end{table}

\begin{table}[h]
    \centering
    \normalsize
    \renewcommand{\arraystretch}{1.2} 
    \begin{tabular}{l|cc}
        \toprule
        & Encoder & Decoder \\
        \midrule
        Num Layers & 6 & 24 \\
        Hidden Size & 768 & 1024 \\
        Num Heads & 12 & 16 \\
        MLP Ratio & 4 & 4 \\
        QKV bias & False & False \\
        MLP bias & False & False \\
        Droppath Rate & 0.0 & 0.1 \\
        Layernorm eps & 1e-6 & 1e-6 \\
        Initializer range & 0.02 & 0.02 \\
        Use Calss Token & False & False \\
        \bottomrule
    \end{tabular}
    \caption{Detail config of ViT Tokenizer.}
    \label{tab: vit configuration}
\end{table}

Our tokenizer adopts the ViT architecture with specific model parameters as shown in the Table \ref{tab: vit configuration}. We process images into tokens using a convolutional kernel with patch size $p = 16$. 

For positional encoding, we employ learnable absolute positional encoding without using rotary positional encoding. To handle multi-scale features at resolutions $s \in \{1,2,4,8,16\}$, we design $h \times w$ positional encodings $PE_{\text{spatial}} \in \mathbb{R}^{H \times W \times d}$, where $H$ and $W$ are the maximum height and width across all scales and 5 additional scale-specific encodings $PE_{\text{scale}} \in \mathbb{R}^{5 \times d}$. For tokens at resolution $s$, we first interpolate the spatial positional encoding
$ PE_{\text{spatial}}$
to target size $(H/s, W/s)$ using area-pooling interpolation, then add the corresponding scale encoding
$ PE_{\text{scale}}^{(s)} \in \mathbb{R}^d, $
resulting in the final positional encoding
$$ PE^{(s)} = \text{Interpolate}(PE_{\text{spatial}}, (H/s, W/s)) + PE_{\text{scale}}^{(s)}. $$

Similar to ViTDet's design, in the multi-scale downsampling module of the convolution, we process each resolution $s$ using three convolutional kernels, downsampling the token map from the encoder's maximum resolution to the corresponding target resolution.

\section{DiT Training and Inference Details}
\label{sec:vit training}
\begin{table}[h]
    \centering
    \normalsize
    \renewcommand{\arraystretch}{1.2} 
    \begin{tabular}{l|cc}
        \toprule
        \textbf{Parameter} & \textbf{Value} \\
        \midrule
        Traing data & ImageNet  \\
        Training steps & 400k  \\
        Global Batch size & 256  \\
        Optimizer & AdamW  \\
        Learning rate & $1.0 \times 10^{-4}$  \\
        LR scheduler & Constant  \\
        beta1 & 0.9 \\
        beta2 & 0.999 \\
        weight decay & 0.0 \\
        \bottomrule
    \end{tabular}
    \caption{Training settings for DiT generation model.}
    \label{tab: config of vit training}
\end{table}
Our primary experiments utilize DiT-XL for training generative models, with detailed parameters illustrated in the accompanying figure. The DiT-XL model requires approximately 40 hours of training on one H20 node to complete 400k steps.

For generative performance evaluation, we perform inference using DiT-XL without classifier-free guidance (CFG). During visualization, we employ a randomly selected CFG scale of 4.2.

\end{document}

%% file: math_commands.tex
\usepackage{amsmath,amsfonts,bm}

\def\eqref#1{equation~\ref{#1}}

\def\1{\bm{1}}

\DeclareMathAlphabet{\mathsfit}{\encodingdefault}{\sfdefault}{m}{sl}
\SetMathAlphabet{\mathsfit}{bold}{\encodingdefault}{\sfdefault}{bx}{n}













%% file: main.bbl
\begin{thebibliography}{64}
\providecommand{\natexlab}[1]{#1}
\providecommand{\url}[1]{\texttt{#1}}
\expandafter\ifx\csname urlstyle\endcsname\relax
  \providecommand{\doi}[1]{doi: #1}\else
  \providecommand{\doi}{doi: \begingroup \urlstyle{rm}\Url}\fi

\bibitem[sd-(2023)]{sd-vae-ft-ema}
stabilityai/sd-vae-ft-ema.
\newblock \url{https://huggingface.co/stabilityai/sd-vae-ft-ema}, 2023.

\bibitem[Bachmann et~al.(2025)Bachmann, Allardice, Mizrahi, Fini, Kar, Amirloo, El-Nouby, Zamir, and Dehghan]{flextok}
Roman Bachmann, Jesse Allardice, David Mizrahi, Enrico Fini, O{\u{g}}uzhan~Fatih Kar, Elmira Amirloo, Alaaeldin El-Nouby, Amir Zamir, and Afshin Dehghan.
\newblock Flextok: Resampling images into 1d token sequences of flexible length.
\newblock In \emph{Forty-second International Conference on Machine Learning}, 2025.

\bibitem[Bank et~al.(2023)Bank, Koenigstein, and Giryes]{ae}
Dor Bank, Noam Koenigstein, and Raja Giryes.
\newblock Autoencoders.
\newblock \emph{Machine learning for data science handbook: data mining and knowledge discovery handbook}, pp.\  353--374, 2023.

\bibitem[Barratt \& Sharma(2018)Barratt and Sharma]{IS}
Shane Barratt and Rishi Sharma.
\newblock A note on the inception score.
\newblock \emph{arXiv preprint arXiv:1801.01973}, 2018.

\bibitem[Caron et~al.(2021)Caron, Touvron, Misra, J{\'e}gou, Mairal, Bojanowski, and Joulin]{dino}
Mathilde Caron, Hugo Touvron, Ishan Misra, Herv{\'e} J{\'e}gou, Julien Mairal, Piotr Bojanowski, and Armand Joulin.
\newblock Emerging properties in self-supervised vision transformers.
\newblock In \emph{Proceedings of the IEEE/CVF international conference on computer vision}, pp.\  9650--9660, 2021.

\bibitem[Chang et~al.(2022)Chang, Zhang, Jiang, Liu, and Freeman]{maskgit}
Huiwen Chang, Han Zhang, Lu~Jiang, Ce~Liu, and William~T Freeman.
\newblock Maskgit: Masked generative image transformer.
\newblock In \emph{Proceedings of the IEEE/CVF conference on computer vision and pattern recognition}, pp.\  11315--11325, 2022.

\bibitem[Chen et~al.(2025)Chen, Han, Chen, Li, Wang, Wang, Wang, Liu, Zou, and Raj]{maetok}
Hao Chen, Yujin Han, Fangyi Chen, Xiang Li, Yidong Wang, Jindong Wang, Ze~Wang, Zicheng Liu, Difan Zou, and Bhiksha Raj.
\newblock Masked autoencoders are effective tokenizers for diffusion models.
\newblock In \emph{Forty-second International Conference on Machine Learning}, 2025.

\bibitem[Chen et~al.(2024)Chen, Cai, Chen, Xie, Yang, Tang, Li, Lu, and Han]{dcae}
Junyu Chen, Han Cai, Junsong Chen, Enze Xie, Shang Yang, Haotian Tang, Muyang Li, Yao Lu, and Song Han.
\newblock Deep compression autoencoder for efficient high-resolution diffusion models.
\newblock \emph{arXiv preprint arXiv:2410.10733}, 2024.

\bibitem[Cheng et~al.(2022)Cheng, Misra, Schwing, Kirillov, and Girdhar]{mask2former}
Bowen Cheng, Ishan Misra, Alexander~G Schwing, Alexander Kirillov, and Rohit Girdhar.
\newblock Masked-attention mask transformer for universal image segmentation.
\newblock In \emph{Proceedings of the IEEE/CVF conference on computer vision and pattern recognition}, pp.\  1290--1299, 2022.

\bibitem[Dosovitskiy et~al.(2020)Dosovitskiy, Beyer, Kolesnikov, Weissenborn, Zhai, Unterthiner, Dehghani, Minderer, Heigold, Gelly, et~al.]{vit}
Alexey Dosovitskiy, Lucas Beyer, Alexander Kolesnikov, Dirk Weissenborn, Xiaohua Zhai, Thomas Unterthiner, Mostafa Dehghani, Matthias Minderer, Georg Heigold, Sylvain Gelly, et~al.
\newblock An image is worth 16x16 words: Transformers for image recognition at scale.
\newblock \emph{arXiv preprint arXiv:2010.11929}, 2020.

\bibitem[Esser et~al.(2021)Esser, Rombach, and Ommer]{vqgan}
Patrick Esser, Robin Rombach, and Bjorn Ommer.
\newblock Taming transformers for high-resolution image synthesis.
\newblock In \emph{Proceedings of the IEEE/CVF conference on computer vision and pattern recognition}, pp.\  12873--12883, 2021.

\bibitem[Fan et~al.(2021)Fan, Xiong, Mangalam, Li, Yan, Malik, and Feichtenhofer]{mvit}
Haoqi Fan, Bo~Xiong, Karttikeya Mangalam, Yanghao Li, Zhicheng Yan, Jitendra Malik, and Christoph Feichtenhofer.
\newblock Multiscale vision transformers.
\newblock In \emph{Proceedings of the IEEE/CVF international conference on computer vision}, pp.\  6824--6835, 2021.

\bibitem[Fan et~al.(2024)Fan, Li, Qin, Li, Sun, Rubinstein, Sun, He, and Tian]{fluid}
Lijie Fan, Tianhong Li, Siyang Qin, Yuanzhen Li, Chen Sun, Michael Rubinstein, Deqing Sun, Kaiming He, and Yonglong Tian.
\newblock Fluid: Scaling autoregressive text-to-image generative models with continuous tokens.
\newblock \emph{arXiv preprint arXiv:2410.13863}, 2024.

\bibitem[Gao et~al.(2023)Gao, Zhou, Cheng, and Yan]{mdtv2}
Shanghua Gao, Pan Zhou, Ming-Ming Cheng, and Shuicheng Yan.
\newblock Mdtv2: Masked diffusion transformer is a strong image synthesizer.
\newblock \emph{arXiv preprint arXiv:2303.14389}, 2023.

\bibitem[Gong et~al.(2025)Gong, Zou, Zheng, Yu, Chen, Sun, Zhao, Zhou, Ji, Ru, et~al.]{ming-lite}
Biao Gong, Cheng Zou, Dandan Zheng, Hu~Yu, Jingdong Chen, Jianxin Sun, Junbo Zhao, Jun Zhou, Kaixiang Ji, Lixiang Ru, et~al.
\newblock Ming-lite-uni: Advancements in unified architecture for natural multimodal interaction.
\newblock \emph{arXiv preprint arXiv:2505.02471}, 2025.

\bibitem[Goodfellow et~al.(2020)Goodfellow, Pouget-Abadie, Mirza, Xu, Warde-Farley, Ozair, Courville, and Bengio]{gan}
Ian Goodfellow, Jean Pouget-Abadie, Mehdi Mirza, Bing Xu, David Warde-Farley, Sherjil Ozair, Aaron Courville, and Yoshua Bengio.
\newblock Generative adversarial networks.
\newblock \emph{Communications of the ACM}, 63\penalty0 (11):\penalty0 139--144, 2020.

\bibitem[{Google}(2025)]{gemini}
{Google}.
\newblock Experiment with gemini 2.0 flash native image generation, 2025.
\newblock Google Blog, 2025.
\newblock URL \url{https://developers.googleblog.com/en/experiment-with-gemini-20-flash-native-image-generation/}.

\bibitem[Hansen-Estruch et~al.(2025)Hansen-Estruch, Yan, Chung, Zohar, Wang, Hou, Xu, Vishwanath, Vajda, and Chen]{scaling_vit}
Philippe Hansen-Estruch, David Yan, Ching-Yao Chung, Orr Zohar, Jialiang Wang, Tingbo Hou, Tao Xu, Sriram Vishwanath, Peter Vajda, and Xinlei Chen.
\newblock Learnings from scaling visual tokenizers for reconstruction and generation.
\newblock \emph{arXiv preprint arXiv:2501.09755}, 2025.

\bibitem[He et~al.(2022)He, Chen, Xie, Li, Doll{\'a}r, and Girshick]{mae}
Kaiming He, Xinlei Chen, Saining Xie, Yanghao Li, Piotr Doll{\'a}r, and Ross Girshick.
\newblock Masked autoencoders are scalable vision learners.
\newblock In \emph{Proceedings of the IEEE/CVF conference on computer vision and pattern recognition}, pp.\  16000--16009, 2022.

\bibitem[Heusel et~al.(2017)Heusel, Ramsauer, Unterthiner, Nessler, and Hochreiter]{fid}
Martin Heusel, Hubert Ramsauer, Thomas Unterthiner, Bernhard Nessler, and Sepp Hochreiter.
\newblock Gans trained by a two time-scale update rule converge to a local nash equilibrium.
\newblock \emph{Advances in neural information processing systems}, 30, 2017.

\bibitem[Ho \& Salimans(2022)Ho and Salimans]{cfg}
Jonathan Ho and Tim Salimans.
\newblock Classifier-free diffusion guidance.
\newblock \emph{arXiv preprint arXiv:2207.12598}, 2022.

\bibitem[Ho et~al.(2020)Ho, Jain, and Abbeel]{ddpm}
Jonathan Ho, Ajay Jain, and Pieter Abbeel.
\newblock Denoising diffusion probabilistic models.
\newblock \emph{Advances in neural information processing systems}, 33:\penalty0 6840--6851, 2020.

\bibitem[Hore \& Ziou(2010)Hore and Ziou]{psnr}
Alain Hore and Djemel Ziou.
\newblock Image quality metrics: Psnr vs. ssim.
\newblock In \emph{2010 20th international conference on pattern recognition}, pp.\  2366--2369. IEEE, 2010.

\bibitem[Huang et~al.(2024)Huang, Ji, Gong, Qing, Zhang, Zheng, Wang, Chen, and Yang]{chain-of-sight}
Ziyuan Huang, Kaixiang Ji, Biao Gong, Zhiwu Qing, Qinglong Zhang, Kecheng Zheng, Jian Wang, Jingdong Chen, and Ming Yang.
\newblock Accelerating pre-training of multimodal llms via chain-of-sight.
\newblock \emph{Advances in Neural Information Processing Systems}, 37:\penalty0 75668--75691, 2024.

\bibitem[Karras et~al.(2019)Karras, Laine, and Aila]{stylegan}
Tero Karras, Samuli Laine, and Timo Aila.
\newblock A style-based generator architecture for generative adversarial networks.
\newblock In \emph{Proceedings of the IEEE/CVF conference on computer vision and pattern recognition}, pp.\  4401--4410, 2019.

\bibitem[Kingma et~al.(2013)Kingma, Welling, et~al.]{vae}
Diederik~P Kingma, Max Welling, et~al.
\newblock Auto-encoding variational bayes, 2013.

\bibitem[Kingma et~al.(2019)Kingma, Welling, et~al.]{vae_intro}
Diederik~P Kingma, Max Welling, et~al.
\newblock An introduction to variational autoencoders.
\newblock \emph{Foundations and Trends{\textregistered} in Machine Learning}, 12\penalty0 (4):\penalty0 307--392, 2019.

\bibitem[Li et~al.(2024)Li, Tian, Li, Deng, and He]{mar}
Tianhong Li, Yonglong Tian, He~Li, Mingyang Deng, and Kaiming He.
\newblock Autoregressive image generation without vector quantization.
\newblock \emph{Advances in Neural Information Processing Systems}, 37:\penalty0 56424--56445, 2024.

\bibitem[Li et~al.(2022)Li, Mao, Girshick, and He]{vitdet}
Yanghao Li, Hanzi Mao, Ross Girshick, and Kaiming He.
\newblock Exploring plain vision transformer backbones for object detection.
\newblock In \emph{European conference on computer vision}, pp.\  280--296. Springer, 2022.

\bibitem[Lin et~al.(2017)Lin, Doll{\'a}r, Girshick, He, Hariharan, and Belongie]{FPN}
Tsung-Yi Lin, Piotr Doll{\'a}r, Ross Girshick, Kaiming He, Bharath Hariharan, and Serge Belongie.
\newblock Feature pyramid networks for object detection.
\newblock In \emph{Proceedings of the IEEE conference on computer vision and pattern recognition}, pp.\  2117--2125, 2017.

\bibitem[Lipman et~al.(2022)Lipman, Chen, Ben-Hamu, Nickel, and Le]{flowmodel}
Yaron Lipman, Ricky~TQ Chen, Heli Ben-Hamu, Maximilian Nickel, and Matt Le.
\newblock Flow matching for generative modeling.
\newblock \emph{arXiv preprint arXiv:2210.02747}, 2022.

\bibitem[Liu et~al.(2021)Liu, Lin, Cao, Hu, Wei, Zhang, Lin, and Guo]{swintransformer}
Ze~Liu, Yutong Lin, Yue Cao, Han Hu, Yixuan Wei, Zheng Zhang, Stephen Lin, and Baining Guo.
\newblock Swin transformer: Hierarchical vision transformer using shifted windows.
\newblock In \emph{Proceedings of the IEEE/CVF international conference on computer vision}, pp.\  10012--10022, 2021.

\bibitem[Ma et~al.(2024)Ma, Goldstein, Albergo, Boffi, Vanden-Eijnden, and Xie]{sit}
Nanye Ma, Mark Goldstein, Michael~S Albergo, Nicholas~M Boffi, Eric Vanden-Eijnden, and Saining Xie.
\newblock Sit: Exploring flow and diffusion-based generative models with scalable interpolant transformers.
\newblock In \emph{European Conference on Computer Vision}, pp.\  23--40. Springer, 2024.

\bibitem[Michelucci(2022)]{ae_intro}
Umberto Michelucci.
\newblock An introduction to autoencoders.
\newblock \emph{arXiv preprint arXiv:2201.03898}, 2022.

\bibitem[{OpenAI}(2025)]{gpt-4o}
{OpenAI}.
\newblock Introducing 4o image generation.
\newblock OpenAI Blog, 2025.
\newblock URL \url{https://openai.com/index/introducing-4o-image-generation/}.

\bibitem[Oquab et~al.(2023)Oquab, Darcet, Moutakanni, Vo, Szafraniec, Khalidov, Fernandez, Haziza, Massa, El-Nouby, et~al.]{dinov2}
Maxime Oquab, Timoth{\'e}e Darcet, Th{\'e}o Moutakanni, Huy Vo, Marc Szafraniec, Vasil Khalidov, Pierre Fernandez, Daniel Haziza, Francisco Massa, Alaaeldin El-Nouby, et~al.
\newblock Dinov2: Learning robust visual features without supervision.
\newblock \emph{arXiv preprint arXiv:2304.07193}, 2023.

\bibitem[Peebles \& Xie(2023)Peebles and Xie]{dit}
William Peebles and Saining Xie.
\newblock Scalable diffusion models with transformers.
\newblock In \emph{Proceedings of the IEEE/CVF international conference on computer vision}, pp.\  4195--4205, 2023.

\bibitem[Radford et~al.(2021)Radford, Kim, Hallacy, Ramesh, Goh, Agarwal, Sastry, Askell, Mishkin, Clark, et~al.]{clip}
Alec Radford, Jong~Wook Kim, Chris Hallacy, Aditya Ramesh, Gabriel Goh, Sandhini Agarwal, Girish Sastry, Amanda Askell, Pamela Mishkin, Jack Clark, et~al.
\newblock Learning transferable visual models from natural language supervision.
\newblock In \emph{International conference on machine learning}, pp.\  8748--8763. PmLR, 2021.

\bibitem[Ramesh et~al.(2021)Ramesh, Pavlov, Goh, Gray, Voss, Radford, Chen, and Sutskever]{zero_shot}
Aditya Ramesh, Mikhail Pavlov, Gabriel Goh, Scott Gray, Chelsea Voss, Alec Radford, Mark Chen, and Ilya Sutskever.
\newblock Zero-shot text-to-image generation.
\newblock In \emph{International conference on machine learning}, pp.\  8821--8831. Pmlr, 2021.

\bibitem[Ramesh et~al.(2022)Ramesh, Dhariwal, Nichol, Chu, and Chen]{text_to_image}
Aditya Ramesh, Prafulla Dhariwal, Alex Nichol, Casey Chu, and Mark Chen.
\newblock Hierarchical text-conditional image generation with clip latents.
\newblock \emph{arXiv preprint arXiv:2204.06125}, 1\penalty0 (2):\penalty0 3, 2022.

\bibitem[Razavi et~al.(2019)Razavi, Van~den Oord, and Vinyals]{vqvae2}
Ali Razavi, Aaron Van~den Oord, and Oriol Vinyals.
\newblock Generating diverse high-fidelity images with vq-vae-2.
\newblock \emph{Advances in neural information processing systems}, 32, 2019.

\bibitem[Ren et~al.(2024)Ren, Yu, He, Shen, Yuille, and Chen]{flowar}
Sucheng Ren, Qihang Yu, Ju~He, Xiaohui Shen, Alan Yuille, and Liang-Chieh Chen.
\newblock Flowar: Scale-wise autoregressive image generation meets flow matching.
\newblock \emph{arXiv preprint arXiv:2412.15205}, 2024.

\bibitem[Reuss et~al.(2024)Reuss, Ya{\u{g}}murlu, Wenzel, and Lioutikov]{mdt}
Moritz Reuss, {\"O}mer~Erdin{\c{c}} Ya{\u{g}}murlu, Fabian Wenzel, and Rudolf Lioutikov.
\newblock Multimodal diffusion transformer: Learning versatile behavior from multimodal goals.
\newblock \emph{arXiv preprint arXiv:2407.05996}, 2024.

\bibitem[Rombach et~al.(2022)Rombach, Blattmann, Lorenz, Esser, and Ommer]{ldm}
Robin Rombach, Andreas Blattmann, Dominik Lorenz, Patrick Esser, and Bj{\"o}rn Ommer.
\newblock High-resolution image synthesis with latent diffusion models.
\newblock In \emph{Proceedings of the IEEE/CVF conference on computer vision and pattern recognition}, pp.\  10684--10695, 2022.

\bibitem[Ronneberger et~al.(2015)Ronneberger, Fischer, and Brox]{unet}
Olaf Ronneberger, Philipp Fischer, and Thomas Brox.
\newblock U-net: Convolutional networks for biomedical image segmentation.
\newblock In \emph{Medical image computing and computer-assisted intervention--MICCAI 2015: 18th international conference, Munich, Germany, October 5-9, 2015, proceedings, part III 18}, pp.\  234--241. Springer, 2015.

\bibitem[Russakovsky et~al.(2015)Russakovsky, Deng, Su, Krause, Satheesh, Ma, Huang, Karpathy, Khosla, Bernstein, et~al.]{imagenet}
Olga Russakovsky, Jia Deng, Hao Su, Jonathan Krause, Sanjeev Satheesh, Sean Ma, Zhiheng Huang, Andrej Karpathy, Aditya Khosla, Michael Bernstein, et~al.
\newblock Imagenet large scale visual recognition challenge.
\newblock \emph{International journal of computer vision}, 115:\penalty0 211--252, 2015.

\bibitem[Saharia et~al.(2022)Saharia, Chan, Saxena, Li, Whang, Denton, Ghasemipour, Gontijo~Lopes, Karagol~Ayan, Salimans, et~al.]{Imagen}
Chitwan Saharia, William Chan, Saurabh Saxena, Lala Li, Jay Whang, Emily~L Denton, Kamyar Ghasemipour, Raphael Gontijo~Lopes, Burcu Karagol~Ayan, Tim Salimans, et~al.
\newblock Photorealistic text-to-image diffusion models with deep language understanding.
\newblock \emph{Advances in neural information processing systems}, 35:\penalty0 36479--36494, 2022.

\bibitem[Skorokhodov et~al.(2025)Skorokhodov, Girish, Hu, Menapace, Li, Abdal, Tulyakov, and Siarohin]{diffusability}
Ivan Skorokhodov, Sharath Girish, Benran Hu, Willi Menapace, Yanyu Li, Rameen Abdal, Sergey Tulyakov, and Aliaksandr Siarohin.
\newblock Improving the diffusability of autoencoders.
\newblock \emph{arXiv preprint arXiv:2502.14831}, 2025.

\bibitem[Sun et~al.(2024)Sun, Jiang, Chen, Zhang, Peng, Luo, and Yuan]{llamagen}
Peize Sun, Yi~Jiang, Shoufa Chen, Shilong Zhang, Bingyue Peng, Ping Luo, and Zehuan Yuan.
\newblock Autoregressive model beats diffusion: Llama for scalable image generation.
\newblock \emph{arXiv preprint arXiv:2406.06525}, 2024.

\bibitem[Tian et~al.(2024)Tian, Jiang, Yuan, Peng, and Wang]{var}
Keyu Tian, Yi~Jiang, Zehuan Yuan, Bingyue Peng, and Liwei Wang.
\newblock Visual autoregressive modeling: Scalable image generation via next-scale prediction.
\newblock \emph{Advances in neural information processing systems}, 37:\penalty0 84839--84865, 2024.

\bibitem[Van Den~Oord et~al.(2017)Van Den~Oord, Vinyals, et~al.]{vqvae}
Aaron Van Den~Oord, Oriol Vinyals, et~al.
\newblock Neural discrete representation learning.
\newblock \emph{Advances in neural information processing systems}, 30, 2017.

\bibitem[Van~der Maaten \& Hinton(2008)Van~der Maaten and Hinton]{tsne}
Laurens Van~der Maaten and Geoffrey Hinton.
\newblock Visualizing data using t-sne.
\newblock \emph{Journal of machine learning research}, 9\penalty0 (11), 2008.

\bibitem[Vaswani et~al.(2017)Vaswani, Shazeer, Parmar, Uszkoreit, Jones, Gomez, Kaiser, and Polosukhin]{attention}
Ashish Vaswani, Noam Shazeer, Niki Parmar, Jakob Uszkoreit, Llion Jones, Aidan~N Gomez, {\L}ukasz Kaiser, and Illia Polosukhin.
\newblock Attention is all you need.
\newblock \emph{Advances in neural information processing systems}, 30, 2017.

\bibitem[Wang et~al.(2025)Wang, Tian, Wang, Zhang, Huang, Wu, and Jiang]{simplear}
Junke Wang, Zhi Tian, Xun Wang, Xinyu Zhang, Weilin Huang, Zuxuan Wu, and Yu-Gang Jiang.
\newblock Simplear: Pushing the frontier of autoregressive visual generation through pretraining, sft, and rl.
\newblock \emph{arXiv preprint arXiv:2504.11455}, 2025.

\bibitem[Wang et~al.(2004)Wang, Bovik, Sheikh, and Simoncelli]{ssim}
Zhou Wang, Alan~C Bovik, Hamid~R Sheikh, and Eero~P Simoncelli.
\newblock Image quality assessment: from error visibility to structural similarity.
\newblock \emph{IEEE transactions on image processing}, 13\penalty0 (4):\penalty0 600--612, 2004.

\bibitem[Xiong et~al.(2025)Xiong, Liew, Huang, Feng, and Liu]{gigatok}
Tianwei Xiong, Jun~Hao Liew, Zilong Huang, Jiashi Feng, and Xihui Liu.
\newblock Gigatok: Scaling visual tokenizers to 3 billion parameters for autoregressive image generation.
\newblock \emph{arXiv preprint arXiv:2504.08736}, 2025.

\bibitem[Yang et~al.(2024)Yang, Liu, Deng, Kim, Mei, Shen, and Chen]{flux}
Chenglin Yang, Celong Liu, Xueqing Deng, Dongwon Kim, Xing Mei, Xiaohui Shen, and Liang-Chieh Chen.
\newblock 1.58-bit flux.
\newblock \emph{arXiv preprint arXiv:2412.18653}, 2024.

\bibitem[Yao et~al.(2025)Yao, Yang, and Wang]{vavae}
Jingfeng Yao, Bin Yang, and Xinggang Wang.
\newblock Reconstruction vs. generation: Taming optimization dilemma in latent diffusion models.
\newblock \emph{arXiv preprint arXiv:2501.01423}, 2025.

\bibitem[Yu et~al.(2023)Yu, Lezama, Gundavarapu, Versari, Sohn, Minnen, Cheng, Birodkar, Gupta, Gu, et~al.]{mgvit2}
Lijun Yu, Jos{\'e} Lezama, Nitesh~B Gundavarapu, Luca Versari, Kihyuk Sohn, David Minnen, Yong Cheng, Vighnesh Birodkar, Agrim Gupta, Xiuye Gu, et~al.
\newblock Language model beats diffusion--tokenizer is key to visual generation.
\newblock \emph{arXiv preprint arXiv:2310.05737}, 2023.

\bibitem[Yu et~al.(2024{\natexlab{a}})Yu, Weber, Deng, Shen, Cremers, and Chen]{titok}
Qihang Yu, Mark Weber, Xueqing Deng, Xiaohui Shen, Daniel Cremers, and Liang-Chieh Chen.
\newblock An image is worth 32 tokens for reconstruction and generation.
\newblock \emph{Advances in Neural Information Processing Systems}, 37:\penalty0 128940--128966, 2024{\natexlab{a}}.

\bibitem[Yu et~al.(2024{\natexlab{b}})Yu, Kwak, Jang, Jeong, Huang, Shin, and Xie]{repa}
Sihyun Yu, Sangkyung Kwak, Huiwon Jang, Jongheon Jeong, Jonathan Huang, Jinwoo Shin, and Saining Xie.
\newblock Representation alignment for generation: Training diffusion transformers is easier than you think.
\newblock \emph{arXiv preprint arXiv:2410.06940}, 2024{\natexlab{b}}.

\bibitem[Zhang et~al.(2018)Zhang, Isola, Efros, Shechtman, and Wang]{lpips}
Richard Zhang, Phillip Isola, Alexei~A Efros, Eli Shechtman, and Oliver Wang.
\newblock The unreasonable effectiveness of deep features as a perceptual metric.
\newblock In \emph{Proceedings of the IEEE conference on computer vision and pattern recognition}, pp.\  586--595, 2018.

\bibitem[Zheng et~al.(2022)Zheng, Vuong, Cai, and Phung]{movqgan}
Chuanxia Zheng, Tung-Long Vuong, Jianfei Cai, and Dinh Phung.
\newblock Movq: Modulating quantized vectors for high-fidelity image generation.
\newblock \emph{Advances in Neural Information Processing Systems}, 35:\penalty0 23412--23425, 2022.

\bibitem[Zheng et~al.(2023)Zheng, Nie, Vahdat, and Anandkumar]{maskdit}
Hongkai Zheng, Weili Nie, Arash Vahdat, and Anima Anandkumar.
\newblock Fast training of diffusion models with masked transformers.
\newblock \emph{arXiv preprint arXiv:2306.09305}, 2023.

\end{thebibliography}
